\documentclass[10pt,twocolumn,letterpaper]{article}

\usepackage{cvpr}
\usepackage{times}
\usepackage{epsfig}
\usepackage{graphicx}
\usepackage{amsmath}
\usepackage{amssymb}
\usepackage{multirow}
\usepackage{rotating}

\usepackage{appendix}

\usepackage[pagebackref=true,breaklinks=true,letterpaper=true,colorlinks,bookmarks=false]{hyperref}

\cvprfinalcopy 

\def\cvprPaperID{3324} 

\ifcvprfinal\fi
\begin{document}

\title{Deep Depth Completion of a Single RGB-D Image}

\vspace{-4mm}
\author{
Yinda Zhang\\
Princeton University
\and
Thomas Funkhouser\\
Princeton University}

\maketitle
\vspace{-2mm}

\begin{abstract}

The goal of our work is to complete the depth channel of an RGB-D image.  Commodity-grade depth cameras often
fail to sense depth for shiny, bright, transparent, and distant surfaces.  
To address this problem, we train a deep network that takes an RGB image as input and
predicts dense surface normals and occlusion boundaries.   Those predictions are then
combined with raw depth observations provided by the RGB-D camera to solve
for depths for all pixels, including those missing in the original observation. This method was
chosen over others (e.g., inpainting depths directly) as the result of 
extensive experiments with a new depth completion benchmark dataset,
where holes are filled in training data through the rendering of 
surface reconstructions created from multiview RGB-D scans.
Experiments with different network inputs, depth representations, 
loss functions, optimization methods, inpainting methods, and 
deep depth estimation networks show that our proposed approach 
provides better depth completions than these alternatives.

\end{abstract}


\vspace*{-5mm}
\section{Introduction}
\label{sec:introduction}

\begin{figure}
\centering
\includegraphics[width=\linewidth]{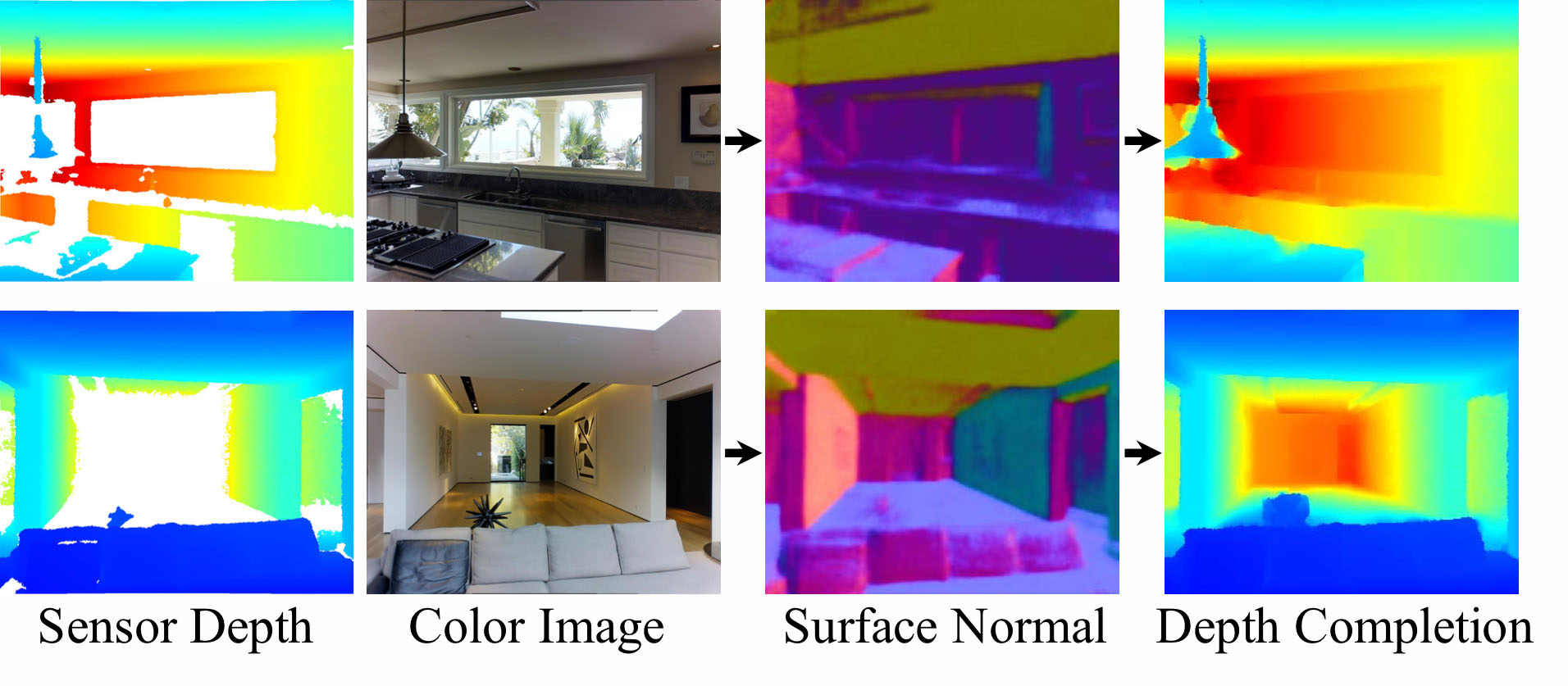}
\caption{{\bf Depth Completion.} We fill in large missing areas in the depth channel of an RGB-D image by predicting normals from color and then solving for completed depths.}
\label{fig:teaser}
\end{figure}

Depth sensing has become pervasive in applications as diverse as
autonomous driving, augmented reality, and scene reconstruction.  
Despite recent advances in depth sensing technology, commodity-level
RGB-D cameras like Microsoft Kinect, Intel RealSense, and Google Tango
still produce depth images with missing data when surfaces are too
glossy, bright, thin, close, or far from the camera.
These problems appear when rooms are large,
surfaces are shiny, and strong lighting is abundant -- 
e.g., in museums, hospitals, 
classrooms, stores, etc.   Even in homes, 
depth images often are missing more than 50\% 
of the pixels (Figure \ref{fig:teaser}).


The goal of our work is to complete
the depth channel of an RGB-D image captured with a commodity camera 
(i.e., fill all the holes).  
Though depth inpainting has received a lot of attention over the past 
two decades, it has generally been addressed with hand-tuned methods
that fill holes by extrapolating boundary surfaces \cite{matsuo2015depth} or 
with Markovian image synthesis \cite{doria2012filling}.  
Newer methods have been proposed to estimate depth de novo from color 
using deep networks \cite{eigen2014depth}.   However, they have not been used for
depth completion, which has its own unique challenges:

\vspace*{1mm}\noindent{\bf Training data:} Large-scale training sets are not
readily available for captured RGB-D images paired with "completed"
depth images (e.g., where ground-truth depth is provided for holes).
As a result, most methods for depth estimation are trained and
evaluated only for pixels that are captured by commodity RGB-D cameras
\cite{silberman2012indoor}.  From this data, they can at-best learn to reproduce
observed depths, but not complete depths that are unobserved, which
have significantly different characteristics.  
To address this issue,
we introduce a new dataset with 105,432 RGB-D images aligned with completed
depth images computed from large-scale surface reconstructions in 72
real-world environments.

\begin{figure*}
\vspace{-4mm}
\centering
\includegraphics[width=\linewidth]{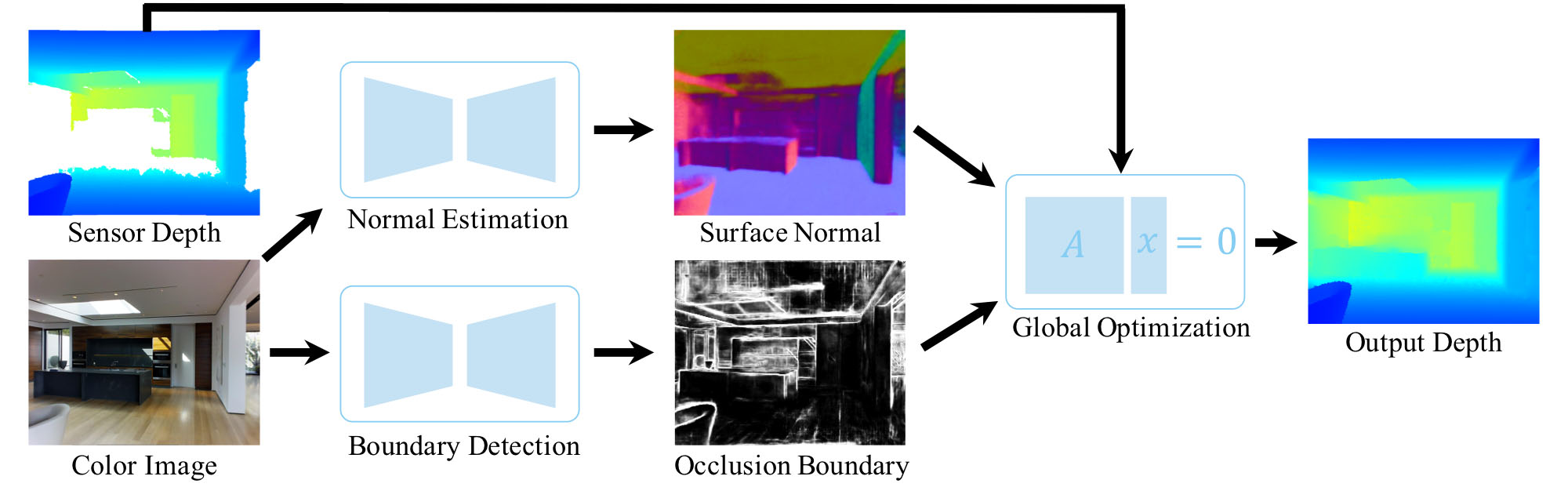}
\caption{{\bf System pipeline.} Given an input RGB-D image, we predict surface normals and occlusion boundaries from color, and then solve for the output depths with a global linear optimization regularized by the input depth.
}
\label{fig:pipeline}
\end{figure*}

\vspace*{1mm}\noindent{\bf Depth representation:} The obvious approach to address
our problem is to use the new dataset as supervision to train a fully
convolutional network to regress depth directly from
RGB-D.  However, that approach does not work very well,
especially for large holes like the one shown in the bottom row of Figure \ref{fig:teaser}.
Estimating absolute depths from a monocular color image
is difficult even for people \cite{mingolla1986perception}.  
Rather, we
train the network to predict only local differential properties of 
depth (surface normals and occlusion boundaries), which are much easier to estimate \cite{koenderink1992surface}.  We then solve for the
absolute depths with a global optimization.

\vspace*{1mm}\noindent{\bf Deep network design:} There is no previous work on
studying how best to design and train an end-to-end deep network for
completing depth images from RGB-D inputs.  At first glance, it seems
straight-forward to extend previous networks trained for
color-to-depth (e.g., by providing them an extra depth channel as
input).  However, we found it difficult to train the networks to
fill large holes from depth inputs -- they generally learn only to copy and interpolate 
the input depth.  It is also challenging for 
the network to learn how to adapt for misalignments of color and depth.
Our solution is to provide
the network with only color images as input (Figure
\ref{fig:pipeline}).  We train it to predict local surface normals
and occlusion boundaries with supervision.  We later combine those
predictions with the input depths in a global optimization to solve back to the completed depth. In this way, the network predicts only local features from color, 
a task where it excels.  
The coarse-scale structure of the scene is reconstructed 
through global optimization with regularization from the input depth.  

\vspace*{1mm} Overall, our main algorithmic insight is that it is best to decompose RGB-D depth completion into two stages: 1) prediction of 
surface normals and occlusion boundaries only from color, 
and 2) optimization of global surface 
structure from those predictions with soft constraints 
provided by observed depths.
During experiments we find with this proposed approach has significantly smaller relative error than alternative approaches.  It has the extra benefit that the 
trained network is independent of the observed depths and so does not need to be retrained for new depth sensors. 



\section{Related Work}
\label{sec:related}

There has been a large amount of prior work on depth estimation, inpainting, and processing.   

\vspace*{1mm}\noindent{\bf Depth estimation.}
Depth estimation from a monocular color image is a long-standing problem in computer vision.
Classic methods include shape-from-shading \cite{zhang1999shape} and shape-from-defocus \cite{suwajanakorn2015depth}.
Other early methods were based on hand-tuned models
and/or assumptions about surface orientations \cite{hoiem2005popup,saxena2006learning,saxena2009make3d}.
Newer methods treat depth estimation as a machine learning problem, most recently using deep networks \cite{eigen2014depth,xie2016deep3d}.
For example, Eigen et al. first used a multiscale convolutional network to regress from 
color images to depths \cite{eigen2014depth,eigen2015predicting}.  
Laina et al. used a fully convolutional network architecture based on ResNet \cite{laina2016deeper}.
Liu et al. proposed a deep convolutional neural field model combining deep networks with Markov random fields \cite{liu2016learning}.
Roy et al. combined shallow convolutional networks with regression forests to reduce the need for large training sets \cite{roy2016monocular}.
All of these methods are trained only to reproduce the raw depth acquired with commodity RGB-D cameras.  
In contrast, we focus on depth completion, where the explicit
goal is to make novel predictions for pixels where the depth sensor has no return.  Since these
pixels are often missing in the raw depth, methods trained only
on raw depth as supervision do not predict them well.

\vspace*{1mm}\noindent{\bf Depth inpainting.}
Many methods have been proposed for filling holes in depth channels of RGB-D images,
including ones that employ smoothness priors \cite{herrera2013depth},  fast marching methods \cite{gong2013guided,liu2012guided}, Navier-Stokes \cite{bertalmio2001navier}, anisotropic diffusion \cite{liu2013guided}, background surface extrapolation \cite{matsuo2015depth,muddala2014depth,thabet20143d}, color-depth edge alignment \cite{chen2014improved,zhang2017probability,zuo2016explicit}, low-rank matrix completion \cite{xue2017depth}, tensor voting \cite{kulkarni2013depth}, Mumford-Shah functional optimization \cite{liu2016building}, joint optimization with other properties of intrinsic images \cite{barron2013intrinsic},
and patch-based image synthesis \cite{ciotta2016depth,doria2012filling,gautier2011depth}.
Recently, methods have been proposed for inpainting {\em color} images with auto-encoders \cite{van2016conditional} and GAN architectures \cite{pathak2016context}. 
However, prior work has not 
investigated how to use those methods for 
inpainting of depth images.  This problem is more difficult due to the
absence of strong features in depth images and the lack 
of large training datasets, an issue addressed in this paper.

\vspace*{1mm}\noindent{\bf Depth super-resolution.}
Several methods have been proposed to improve the spatial resolution of depth images using high-resolution color.   
They have exploited a variety of approaches, including Markov random fields \cite{mac2012patch,diebel2006application,lu2011revisit,park2011high,shabaninia2017high}, shape-from-shading \cite{han2013high,yu2013shading}, segmentation \cite{lu2015sparse}, and dictionary methods \cite{freeman2002example,kiechle2013joint,mahmoudi2012sparse,tosic2014learning}.  
Although some of these techniques may be used for depth completion, the challenges of super-resolution are quite different -- there the focus is on improving spatial resolution, where low-resolution measurements are assumed to be complete and regularly sampled.  In contrast, our focus is on filling holes, which can be quite large and complex and thus require synthesis of large-scale content.

\vspace*{1mm}\noindent{\bf Depth reconstruction from sparse samples.}
Other work has investigated depth reconstruction from color images augmented 
with sparse sets of depth measurements.
Hawe et al. investigated using a Wavelet basis for reconstruction \cite{hawe2011dense}.  
Liu et al. combined wavelet and contourlet dictionaries \cite{liu2015depth}.  
Ma et al. showed that providing $\sim$100 well-spaced depth samples improves depth estimation over color-only methods by two-fold for NYUv2 \cite{ma2017sparse}, yet still with relatively low-quality results.   These methods share some ideas with our work.   However, their motivation is to reduce the cost of sensing in specialized settings (e.g., to save power on a robot), not to complete data typically missed in readily available depth cameras.


\section{Method}
\label{sec:method}

In this paper, we investigate how to use a deep network to complete the depth channel of a single RGB-D image.   Our investigation focuses on the following questions: ``how can we get training data for depth completion?,'' ``what depth representation should we use?,'' and ``how should cues from color and depth be combined?.'' 




\subsection{Dataset}
\label{sec:dataset}

The first issue we address is to create a dataset of RGB-D images paired with completed depth images. 

A straight-forward approach to this task would be to capture images with a low-cost RGB-D camera and align them to images captured simultaneously with a higher cost depth sensor.  This approach is costly and time-consuming -- the largest public datasets of this type cover a handful of indoor scenes (e.g., \cite{park2017colored,scharstein2014high,xue2017depth}).  

Instead, to create our dataset, we utilize existing surface meshes reconstructed from multi-view RGB-D scans of large environments.   There are several datasets of this type, including Matterport3D \cite{chang2017matterport3D}, ScanNet \cite{dai2017scannet}, SceneNN \cite{hua2016scenenn}, and SUN3D \cite{halber2017fine,xiao2013sun3d}, to name a few.   We use Matterport3D.  For each scene, we extract a triangle mesh $M$ with $\sim$1-6 million triangles per room from a global surface reconstruction using screened Poisson surface reconstruction \cite{kazhdan2013screened}.  Then, for a sampling of RGB-D images in the scene, we render the reconstructed mesh $M$ from the camera pose of the image viewpoint to acquire a completed depth image D*.  This process provides us with a set of RGB-D $\rightarrow$ D* image pairs without having to collect new data.  

\begin{figure}
\centering
\includegraphics[width=\linewidth]{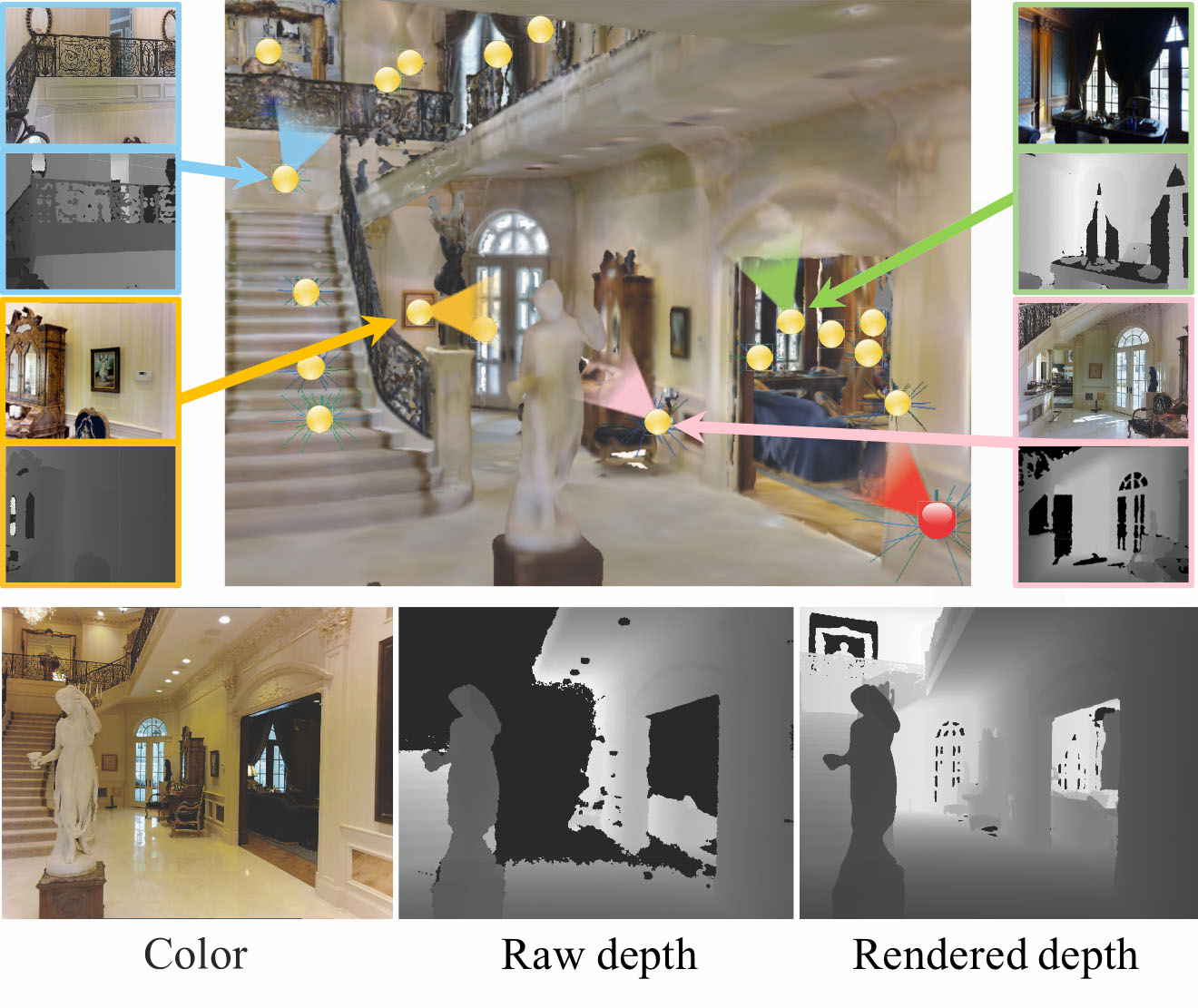}
\vspace{-5mm}
\caption{{\bf Depth Completion Dataset.}  Depth completions are computed
from multi-view surface reconstructions of large indoor environments.
In this example, the bottom shows the raw color and depth channels with the rendered depth for the viewpoint marked as the red dot.  
The rendered mesh (colored by vertex in large image) is created by 
combining RGB-D images 
from a variety of other views spread throughout the scene (yellow dots), 
which collaborate to fill holes when rendered to the red dot view.
}
\label{fig:dataset}
\end{figure}

Figure \ref{fig:dataset} shows some examples of depth image completions from our dataset.  Though the completions are not always perfect, they have several favorable properties for training a deep network for our problem \cite{meister2012can}.
First, the completed depth images generally have fewer holes.  
That's because it is not limited by the observation of one camera viewpoint (e.g., the red dot in Figure \ref{fig:dataset}), but instead by the union of all observations of all cameras viewpoints contributing to the surface reconstruction (yellow dots in Figure \ref{fig:dataset}).   As a result, surfaces distant to one view, but within range of another, will be included in the completed depth image.   Similarly, glossy surfaces that provide no depth data when viewed at a grazing angle usually can be filled in with data from other cameras viewing the surface more directly (note the completion of the shiny floor in rendered depth).  On average, 64.6\% of the pixels missing from the raw depth images are filled in by our reconstruction process.

Second, the completed depth images generally replicate the resolution of the originals for close-up surfaces, but provide far better resolution for distant surfaces.  Since the surface reconstructions are constructed at a 3D grid size comparable to the resolution of a depth camera, there is usually no loss of resolution in completed depth images.   However, that same 3D resolution provides an effectively higher pixel resolution for surfaces further from the camera when projected onto the view plane.  As a result, completed depth images can leverage subpixel antialiasing when rendering high resolution meshes to get finer resolution than the originals (note the detail in the furniture in Figure \ref{fig:dataset}).

Finally, the completed depth images generally have far less noise than the originals.   Since the surface reconstruction algorithm combines noisy depth samples from many camera views by filtering and averaging, it essentially de-noises the surfaces.   This is especially important for distant observations (e.g., $>$4 meters), where raw depth measurements are quantized and noisy.  

In all, our dataset contains 117,516 RGB-D images with rendered completions, which we split into a training set with 105,432 images and a test set with 12,084 images.


\subsection{Depth Representation}
\label{sec:representation}

A second interesting question is ``what geometric representation is best for deep depth completion?''   

A straight-forward approach is to design a network that regresses completed depth from raw depth and color.   However, absolute depth can be difficult to predict from monocular images,
as it may require knowledge of object sizes, scene categories, etc.  Instead, we train the network to predict local properties of the visible surface at each pixel and then solve back for the depth from those predictions.  

Previous work has considered a number of indirect representations of depth.  For example, Chen et al. investigated relative depths \cite{chen2016single}.  Charkrabarti et al. proposed depth derivatives  \cite{chakrabarti2016depth}.  Li et al. used depth derivatives in conjunction with depths \cite{li2017two}.    We have experimented with methods based on predicted derivatives.   However, we find that they do not perform the best in our experiments (see Section \ref{sec:results}). 

Instead, we focus on predicting surface normals and occlusion boundaries.  Since normals are differential surface properties, they depend only on local neighborhoods of pixels.  Moreover, they relate strongly to local lighting variations directly observable in a color image.  For these reasons, previous works on dense prediction of surface normals from color images produce excellent results \cite{bansal2016marr,eigen2015predicting,li2015depth,wang2015designing,zhang2017physically}.  Similarly, occlusion boundaries produce local patterns in pixels (e.g., edges), and so they usually can be robustly detected with a deep network \cite{ehinger2017local,zhang2017physically}.

A critical question, though, is how we can use predicted surface normals 
and occlusion boundaries to complete depth images.  
Several researchers have used predicted normals to refine details on observed 3D surfaces  \cite{hane2015direction,nehab2005efficiently,xie20173D}, and Galliani \etal \cite{galliani2016just} used surface normals to recover missing geometry in multi-view reconstruction for table-top objects.  However, nobody has ever used surface normals before for depth estimation or completion from monocular RGB-D images in complex environments.

\begin{figure}[t]
\centering
\includegraphics[width=\linewidth]{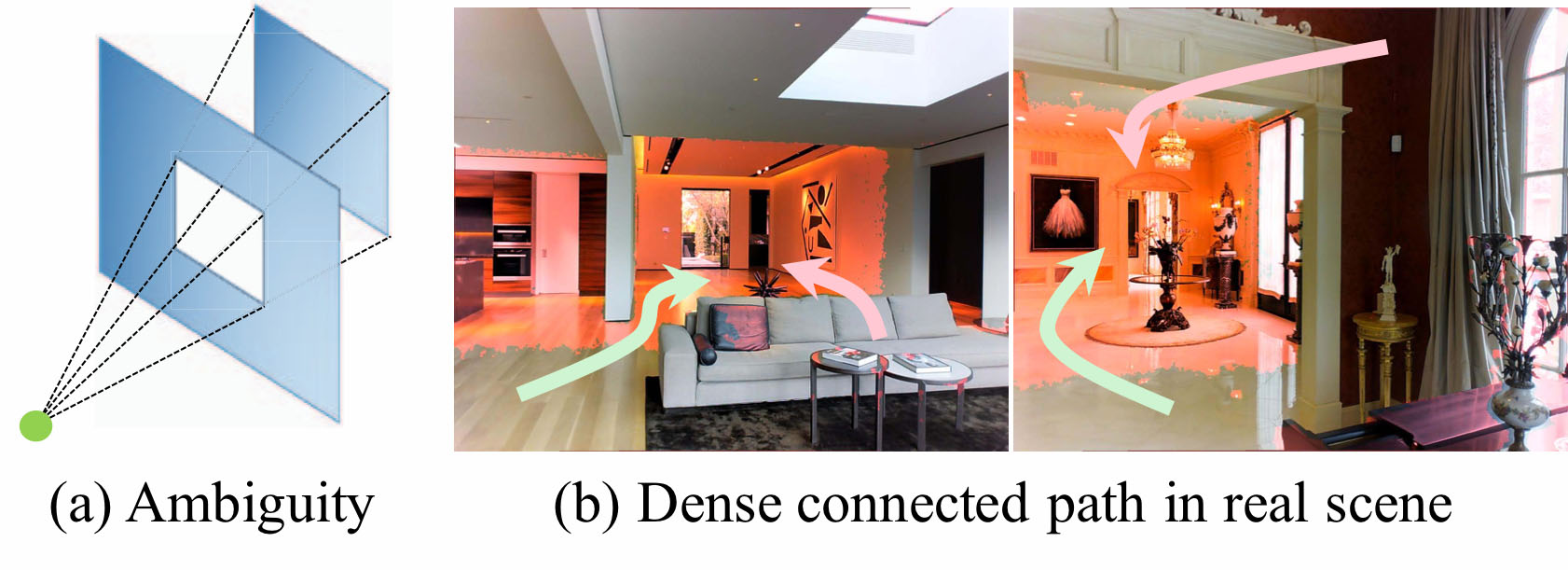}
\caption{{\bf Using surface normals to solve for depth completion.} (a) An example of where depth cannot be solved from surface normal. (b) The area missing depth is marked in red. The red arrow shows paths on which depth cannot be integrated from surface normals. However in real-world images, there are usually many paths through connected neighboring pixels (along floors, ceilings, etc.) over which depths can be integrated (green arrows).
}
\label{fig:underspecified}
\end{figure}

Unfortunately, it is theoretically not possible to solve for depths from only surface normals and occlusion boundaries.   There can be pathological situations where the depth relationships between different parts of the image cannot be inferred only from normals.  For example, in Figure \ref{fig:underspecified}(a), it is impossible to infer the depth of the wall seen through the window based on only the given surface normals.  In this case, the visible region of the wall is enclosed completely by occlusion boundaries (contours) from the perspective of the camera, leaving its depth indeterminate with respect to the rest of the image.   

In practice, however, for real-world scenes it is very unlikely that a region of an image will both be surrounded by occlusion boundaries AND contain no raw depth observations at all (Figure \ref{fig:underspecified}(b)).   Therefore, we find it practical to complete even large holes in depth images using predicted surface normals with coherence weighted by predicted occlusion boundaries and regularization constrained by observed raw depths.   During experiments, we find that solving depth from predicted surface normals and occlusion boundaries results in better depth completions than predicting absolute depths directory, or even solving from depth derivatives (see Section \ref{sec:results}). 

\subsection{Network Architecture and Training}
\label{sec:training}
A third interesting question is ``what is the best way to train a deep network to predict surface normals and occlusion boundaries for depth completion?''

For our study, we pick the deep network architecture proposed in Zhang et.al 
because it has shown competitive performance on both normal estimation and boundary detection \cite{zhang2017physically}.
The model is a fully convolutional neural network built on the back-bone of VGG-16 with symmetry encoder and decoder.
It is also equipped with short-cut connections and shared pooling masks for corresponding max pooling and unpooling layers, 
which are critical for learning local image features.   We train the network with ``ground truth'' surface normals and silhouette boundaries computed from the reconstructed mesh.

After choosing this network, there are still several interesting questions regarding how to training it for depth completion.  
The following paragraphs consider these questions with a focus  
on normal estimation, but the issues and conclusions apply similarly
for occlusion boundary detection. 




\vspace*{2mm}\noindent{\bf What loss should be used to train the network?}
Unlike past work on surface normal estimation, our primary goal is to train a network to predict normals 
{\em only for pixels inside holes} of raw observed depth images.   Since the color appearance characteristics of those pixels 
are likely different than the others (shiny, far from the camera, etc.), one might think that the network should be
supervised to regress normals only for these pixels.   Yet, there are fewer pixels in holes than not, and so training
data of that type is limited.   It was not obvious whether it is best
to train only on holes vs. all pixels.   So, we tested both and compared.  

We define the observed pixels as the ones with depth data from both the raw sensor and the rendered mesh, and the unobserved pixels as the ones with depth from the rendered mesh but not the raw sensor.   For any given set of pixels (observed, unobserved, or both), we train models with a loss for only those pixels by masking out the gradients on other pixels during the back-propagation.


Qualitative and quantitative results comparing the results for different trained models are shown in supplemental material.  The results suggest that the models trained with all pixels perform better than the ones using only observed or only unobserved pixels, and ones trained with rendered normals perform better than with raw normals.

\vspace*{2mm}\noindent{\bf What image channels should be input to the network?}
One might think that the best way to train the network to predict surface normals from a raw RGB-D image is to provide all four channels (RGBD) and train it to regress the three normal channels.   However, surprisingly, we find that our networks performed poorly at predicting normals for pixels without observed depth when trained that way.  They are excellent at predicting normals for pixels with observed depth, but not for the ones in holes -- i.e., the ones required for depth completion.  This result holds regardless of what pixels are included in the loss.   

\begin{figure}
\centering
\includegraphics[width=\linewidth]{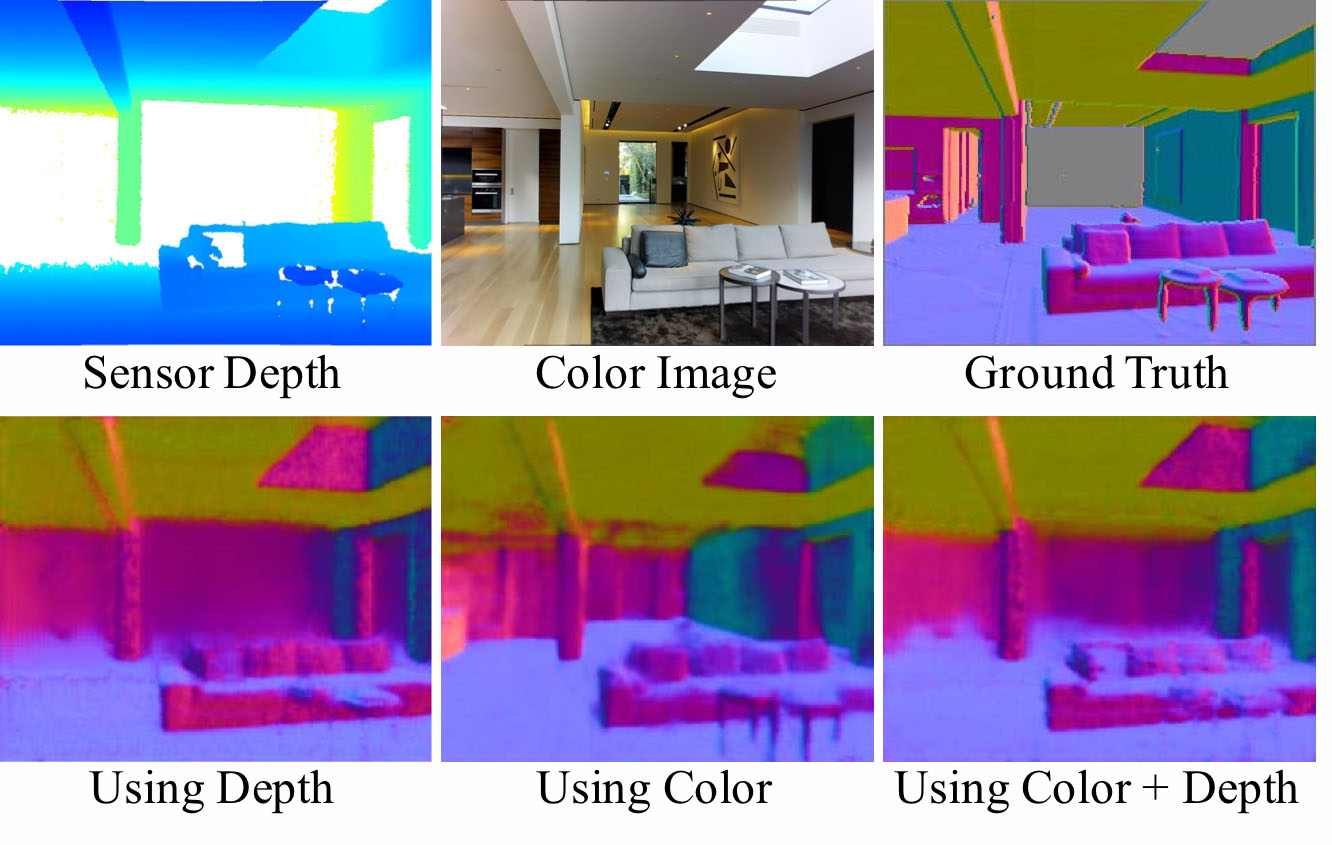}
\caption{{\bf Surface normal estimation for different inputs.} The top row shows
an input color image, raw depth, and the rendered normal. The bottom row shows
surface normal predictions when the inputs are depth only, color only, and both.
The middle one performs the best for the missing area, while comparable elsewhere with the other two models even without depth as input.
}
\label{fig:input_result}
\end{figure}

We conjecture that the network trained with raw depth mainly learns to compute normals from depth directly -- it fails to learn how to predict normals from color when depth is not present, which is the key skill for depth completion.  In general, we find that the network learns to predict normals better from color than depth, even if the network is given an extra channel containing a binary mask indicating which pixels have observed depth \cite{zhang2017real}.  For example, in Figure \ref{fig:input_result}, we see that the normals predicted in large holes from color alone are better than from depth, and just as good as from both color and depth.  Quantitative experiments support this finding in Table \ref{tab:input_results}.

This result is very interesting because it suggests that we can train a network to predict surface normals from color alone and use the observed depth {\em only as regularization} when solving back for depth from normals (next section).   This strategy of separating ``prediction without depth'' from ``optimization with depth'' is compelling for two reasons.   First, the prediction network does not have to be retrained for different depth sensors.  Second, the optimization can be generalized to take a variety of depth observations as regularization, including perhaps sparse depth samples \cite{ma2017sparse}.   This is investigated experimentally in Section \ref{sec:results}.








\begin{table*}
\vspace{-2mm}
\centering
\begin{tabular}{|c|cc|ccccc|cc|ccc|}
\hline
 & \multicolumn{7}{|c|}{Depth Completion} & \multicolumn{5}{|c|}{Surface Normal Estimation} \tabularnewline
\hline
Input &  Rel$\downarrow$ & RMSE$\downarrow$ & 1.05$\uparrow$ & 1.10$\uparrow$ & 1.25$\uparrow$ & $1.25^2$$\uparrow$ & $1.25^3$$\uparrow$ & Mean$\downarrow$ & Median$\downarrow$ & 11.25$\uparrow$ & 22.5$\uparrow$ & 30$\uparrow$ \tabularnewline
\hline 
Depth  & 0.107 & 0.165 & 38.89 & 48.54 & 61.12 & 73.57 & 80.98 & 35.08 & 23.07 & 27.6 & 49.1 & 58.6 \tabularnewline
Both   & 0.090 & 0.124 & 40.13 & 51.26 & 64.84 & 76.46 & 83.05 & 35.30 & 23.59 & 26.7 & 48.5 & 58.1\tabularnewline
Color  & 0.089 & 0.116 & 40.63 & 51.21 & 65.35 & 76.64 & 82.98 & 31.13 & 17.28 & 37.7 & 58.3 & 67.1 \tabularnewline
\hline
\end{tabular}
\vspace{1mm}
\caption{{\bf Effect of different inputs to our deep network.} We train models taking depth, color, and both respectively for surface normal estimation and depth completion. Using only color as input achieves similar performance as the case with both. }
\label{tab:input_results}
\end{table*}

\subsection{Optimization}
\label{sec:optimization}

After predicting the surface normal image $N$ and occlusion boundary image $B$, we solve a system of equations to complete the depth image $D$.
The objective function is defined as the weighted sum of squared errors with four terms:
\begin{small}
\vspace{-1mm}
\begin{align}
\begin{split}
 E & = \lambda_D E_{D} + \lambda_S E_{S} + \lambda_N E_{N} B\\
E_{D}  & = \sum\limits_{p \in T_{obs}}||D(p)-D_0(p)||^2 \\
E_{N}  & = \sum\limits_{p,q \in N}||<v(p,q),N(p)>||^2 \\
E_{S}  & = \sum\limits_{p,q \in N}||D(p)-D(q))|^2  
\end{split}
\label{eq:objective}
\end{align}
\end{small}
where
$E_{D}$ measures the distance between the estimated depth $D(p)$ and the observed raw depth $D_0(p)$ at pixel $p$,
$E_{N}$ measures the consistency between the estimated depth and the predicted surface normal $N(p)$,  
$E_{S}$ encourages adjacent pixels to have the same depths.
$B$ $\in [0,1]$ down-weights the normal terms based on the
predicted probability a pixel is on an occlusion boundary (B(p)).


In its simplest form, this objective function is non-linear, due to the normalization of the tangent vector $v(p,q)$
required for the dot product with the surface normal in $E_{N}$.  However, we can approximate this error term with a
linear formation by foregoing the vector normalization, as suggested in \cite{nehab2005efficiently}.   
In other settings, this approximation would add sensitivity to scaling errors, 
since smaller depths result in shorter tangents and potentially smaller $E_N$ terms.
However, in a depth completion setting, the data term $E_D$ forces the global solution to maintain the correct 
scale by enforcing consistency with the observed raw depth, and thus this is not a significant problem.

Since the matrix form of the system of equations is sparse and symmetric positive definite, we can 
solve it efficiently with a sparse Cholesky factorization (as implemented in cs\_cholsol in CSparse \cite{davis2006csparse}).   
The final solution is a global minimum to the approximated objective function.   

This linearization approach is critical to the success of the proposed method.   Surface normals and occlusion boundaries (and optionally depth derivatives) capture only local properties of the surface geometry, which makes them relatively easy to estimate.   
Only through global optimization can we combine them to complete the depths for all pixels in a consistent solution.


\section{Experimental Results}
\label{sec:results}

We ran a series of experiments to test the proposed methods.   
Unless otherwise specified, 
networks were pretrained on the 
SUNCG dataset \cite{song2016semantic,zhang2017physically} and
fine-tuned on the training split of the our new dataset 
using only color as input and a loss computed for all rendered pixels.  
Optimizations were performed with 
$\lambda_D = 10^3$, $\lambda_N = 1$, and $\lambda_S = 10^{-3}$.
Evaluations were performed on the test split of our new dataset.

We find that predicting surface normals and occlusion boundaries from color at 320x256 
takes $\sim$0.3 seconds on a NVIDIA TITAN X GPU.  Solving the linear equations for depths 
takes $\sim$1.5 seconds on a Intel Xeon 2.4GHz CPU.


\subsection{Ablation Studies}
\label{sec:ablation}

The first set of experiments investigates how different test inputs,
training data, loss functions, depth representations, and 
optimization methods affect the depth prediction results
(further results can be found in the supplemental material).  

Since the focus of our work is predicting depth where it is unobserved
by a depth sensor, our evaluations measure errors in depth predictions
only for pixels of test images {\em unobserved} in the test depth image 
(but present in the rendered image).   This is the opposite of most 
previous work on depth estimation, where error is measured only for 
pixels that are observed by a depth camera.

When evaluating depth predictions, we report
the median error relative to the rendered depth (Rel),
the root mean squared error in meters (RMSE), and percentages of 
pixels with predicted depths falling within an interval 
([$\delta =|predicted-true|/true]$), where $\delta$ is  
$1.05$, $1.10$, $1.25$, $1.25^{2}$, or $1.25^{3}$.   These metrics
are standard among previous work on depth prediction, except that we add
thresholds of $1.05$ and $1.10$ to enable finer-grained evaluation.

When evaluating surface normal predictions, we report
the mean and median errors (in degrees), plus the percentages of 
pixels with predicted normals less than thresholds of 
11.25, 22.5, and 30 degrees.

\vspace*{2mm}\noindent{\bf What data should be input to the network?}
Table \ref{tab:input_results} shows results of an experiment to test
what type of inputs are best for our normal prediction network: color only, raw depth only,
or both.   Intuitively, it would seem that inputting
both would be best.   However, we find that the network learns to predict surface
normals better when given only color  (median error = 17.28$^{\circ}$ for color vs. 23.07$^{\circ}$ for both), which results in depth estimates that are also slightly better (Rel = 0.089 vs. 0.090).   This difference persists whether we train with depths for all pixels, only observed pixels, or only unobserved pixels (results in supplemental material).   We expect the reason is that the network quickly learns to interpolate from observed depth if it is available, which hinders it from learning to synthesize new depth in large holes.  

The impact of this result is quite significant, as it motivates our two-stage system design that separates normal/boundary prediction only from color and optimization with raw depth.   



\vspace*{2mm}\noindent{\bf What depth representation is best?}
Table \ref{tab:reps} shows results of an experiment to test
which depth representations are best for our network to predict.
We train networks separately to predict absolute depths (D), 
surface normals (N), and depth derivatives in 8 directions (DD),
and then use different combinations to complete the depth by optimizing Equation \ref{eq:objective}.
The results
indicate that solving for depths from predicted normals (N) provides the best 
results (Rel = 0.089 for normals (N) as compared to 0.167 for depth (D), 
0.100 for derivatives (DD), 0.092 for normals and derivatives (N+DD).
We expect that this is because normals represent only the orientation of 
surfaces, which is relatively easy to predict \cite{koenderink1992surface}.  
Moreover, normals do not scale with depth, unlike depths or depth derivatives, and thus are more consistent across a range of views.

\begin{table}[th]
\vspace{-5mm}
\scalebox{0.72}{
\begin{tabular}{|c|c|cc|ccccc|}
\hline 
B & Rep & Rel$\downarrow$ & RMSE$\downarrow$ & 1.05$\uparrow$ & 1.10$\uparrow$ & 1.25$\uparrow$ & $1.25^2$$\uparrow$ & $1.25^3$$\uparrow$ \tabularnewline
\hline 
- & D      & 0.167 & 0.241 & 16.43 & 31.13 & 57.62 & 75.63 & 84.01 \tabularnewline
\hline
\multirow{3}{*}{No} 
 & DD        & 0.123 & 0.176 & 35.39 & 45.88 & 60.41 & 73.26 & 80.73 \tabularnewline
 & N+DD & 0.112 & 0.163 & 37.85 & 47.22 & 61.27 & 73.70 & 80.83 \tabularnewline
 & N      & 0.110 & 0.161 & 38.12 & 47.96 & 61.42 & 73.77 & 80.85 \tabularnewline
\hline
\multirow{3}{*}{Yes} 
 & DD         & 0.100 & 0.131 & 37.95 & 49.14 & 64.26 & 76.14 & 82.63 \tabularnewline
 & N+DD  & 0.092 & 0.122 & 39.93 & 50.73 & 65.33 & \textbf{77.04} & \textbf{83.25} \tabularnewline
 & N        & \textbf{0.089} & \textbf{0.116} & \textbf{40.63} & \textbf{51.21} & \textbf{65.35} & 76.74 & 82.98 \tabularnewline
\hline 
\end{tabular}
}
\vspace{1mm}
\caption{{\bf Effect of predicted representation on depth accuracy.} ``DD'' represents depth derivative, and ``N'' represents surface normal. We also evaluate the effect of using boundary weight. The first row shows the performance of directly estimating depth. Overall, solving back depth with surface normal and occlusion boundary gives the best performance.}
\label{tab:reps}
\vspace{1mm}
\end{table}

\begin{figure}[h]
\centering
\includegraphics[width=\linewidth]{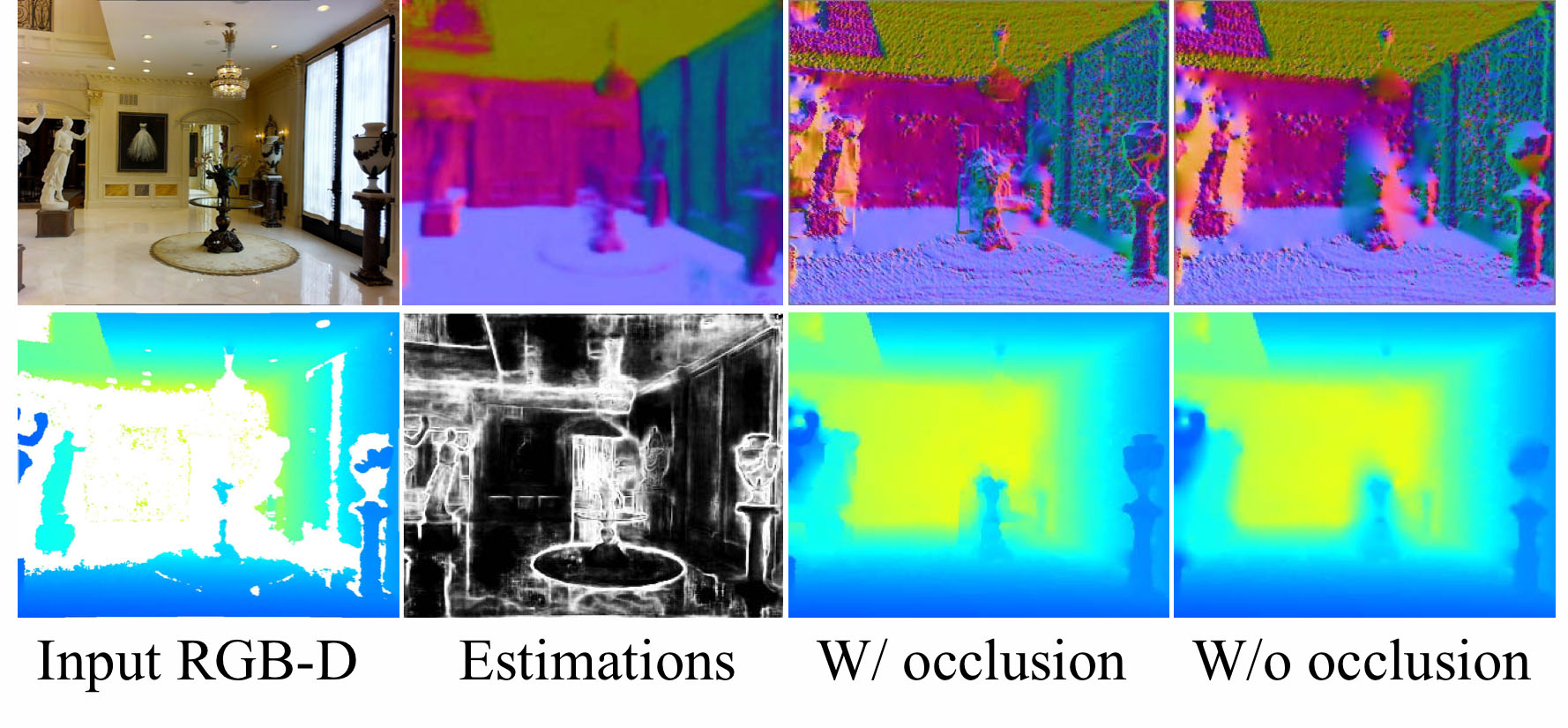}
\vspace{-4mm}
\caption{{\bf Effect of occlusion boundary prediction on normals.} The 2nd column shows the estimated surface normal and occlusion boundary. The 3rd and 4th column shows the output of the optimization with/without occlusion boundary weight. To help understand the 3D geometry and local detail, we also visualize the surface normal computed from the output depth. The occlusion boundary provides information for depth discontinuity, which help to maintain boundary sharpness. }  
\label{fig:boundary_result}
\end{figure}

\vspace*{2mm}\noindent{\bf Does prediction of occlusion boundaries help?}
The last six rows of Table \ref{tab:reps} show results of an experiment to test
whether down-weighting the effect of surface normals near predicted occlusion boundaries
helps the optimizer solve for better depths.  Rows 2-4 are without boundary prediction (``No'' in the first column), and
Rows 5-7 are with (``Yes'').  The results
indicate that boundary predictions 
improve the results by $\sim$19\%  (Rel = 0.089 vs. 0.110).   
This suggests that the network is on average correctly predicting pixels where surface normals are noisy or incorrect, as shown qualitatively in Figure \ref{fig:boundary_result}.

\begin{figure}[t]
\centering
\vspace{-5mm}
\includegraphics[width=\linewidth]{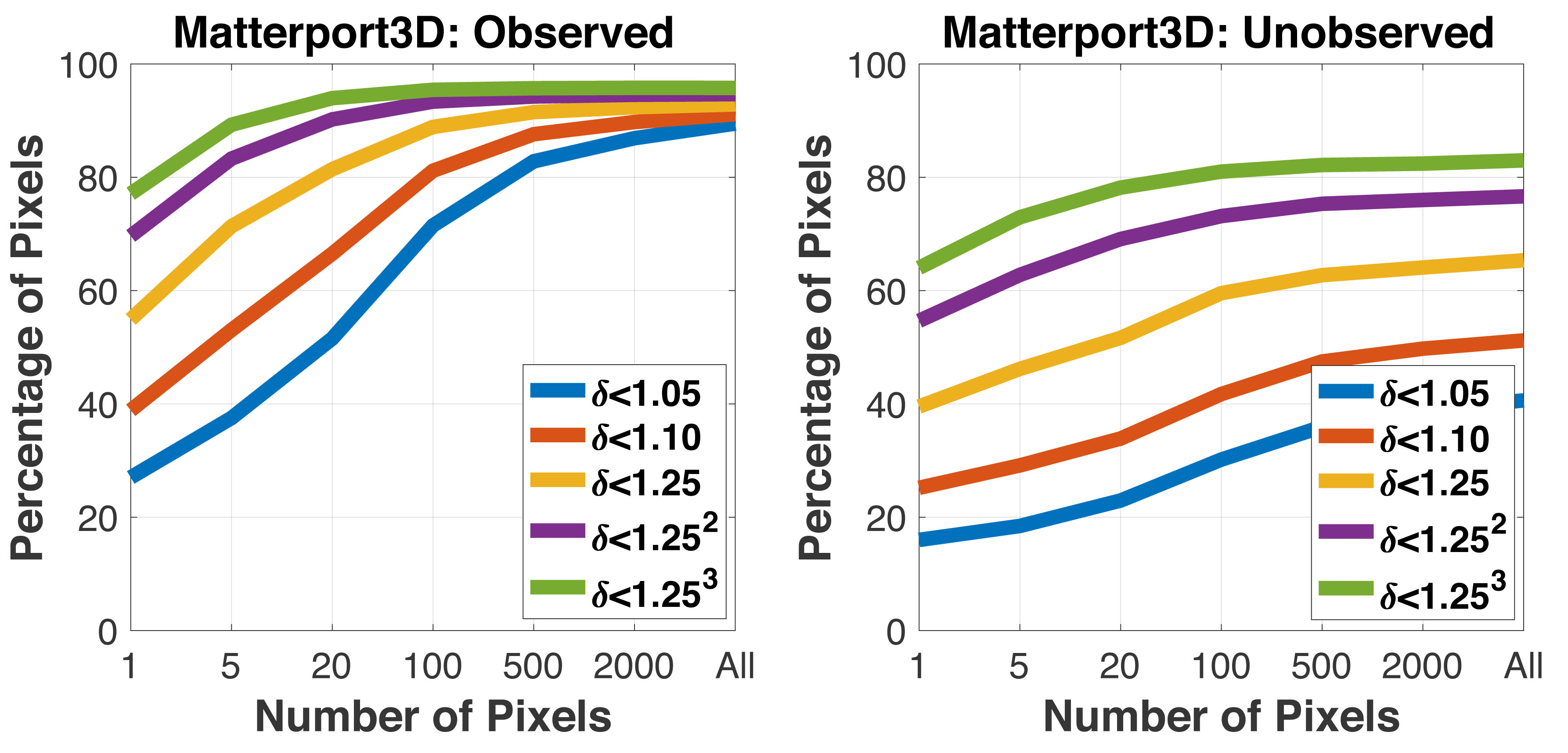}
\caption{{\bf Effect of sparse raw depth inputs on depth accuracy.} The depth completion performance of our method w.r.t number of input pixels with depth. The plot shows that depth estimation on unobserved pixels is harder than the observed. It also shows that our method works well with only a small number of sparse pixels, which is desirable to many applications.}
\label{fig:sparse_results}
\end{figure}

\vspace{-2mm}
\vspace*{2mm}\noindent{\bf How much observed depth is necessary?}
Figure \ref{fig:sparse_results} shows results of an experiment to test how much
our depth completion method depends on the quantity of input depth.   To 
investigate this question, we degraded the input depth
images by randomly masking different numbers
of pixels before giving them to the optimizer
to solve for completed depths from predicted normals and boundaries.
The two plots shows curves indicating depth accuracy solved
for pixels that are observed (left) and unobserved (right) in
the original raw depth images.  From these results, we see that
the optimizer is able to solve for depth almost as accurately 
when given only a small fraction of the pixels in the raw depth image.  
As expected, the performance is
much worse on pixels unobserved by the raw depth (they are harder).
However, the depth estimations are still quite good when only a small fraction of the
raw pixels are provided (the rightmost point on the curve at 2000 
pixels represents only 2.5\% of all pixels).  This results suggests
that our method could be useful for other depth sensor designs with sparse 
measurements.  In this setting, our deep network would not have to be retrained for each new dense sensor (since it depends only on color), a benefit of our two-stage approach.

\subsection{Comparison to Baseline Methods}
\label{sec:comparison}

The second set of experiments investigates how the proposed approach compares to baseline depth inpainting and depth estimation methods.

\begin{figure}
\vspace{-2mm}
\centering
\includegraphics[width=\linewidth]{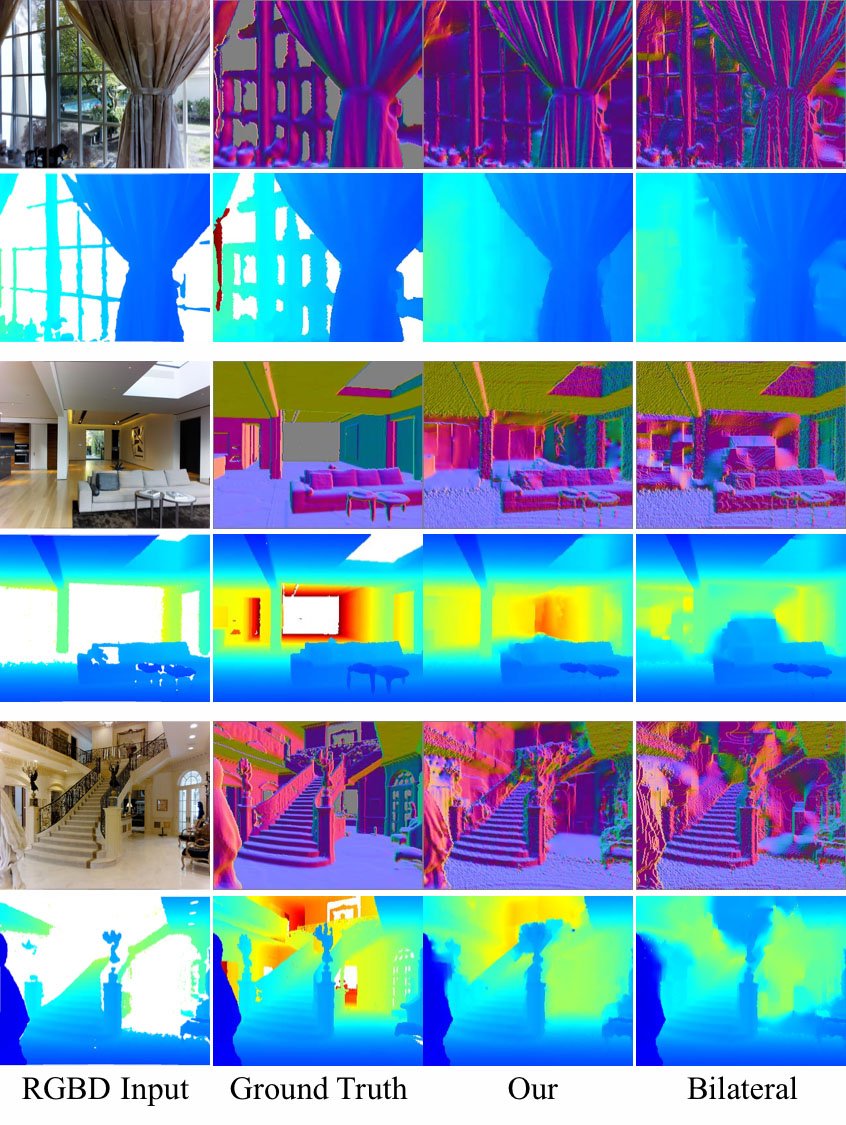}
\caption{{\bf Comparison to inpainting with a joint bilateral filter.} Our method learns better guidance from color and produce comparatively sharper and more accurate results. } 
\label{fig:inpainting_comparsion}
\end{figure}

\vspace*{2mm}\noindent{\bf Comparison to Inpainting Methods}
Table \ref{tab:inpainting_comparison} shows results of a study comparing our proposed method to 
typical non-data-driven alternatives for depth inpainting. 
The focus of this study is to establish 
how well-known methods perform to provide a baseline on how hard the problem is for this new dataset.   As such, the methods we consider include: 
a) joint bilinear filtering \cite{silberman2012indoor} (Bilateral), b) fast bilateral solver \cite{barron2016fast} (Fast), and c) global edge-aware energy optimization \cite{ferstl2013image} (TGV).
The results in Table \ref{tab:inpainting_comparison} show that our method significantly outperforms these methods (Rel=0.089 vs. 0.103-0.151 for the others).  By training to predict surface normals with a deep network, our method learns to complete depth with data-driven priors, which are stronger than simple geometric heuristics.  The difference to the best of the tested hand-tuned approaches (Bilateral) can be seen in Figure \ref{fig:inpainting_comparsion}.

\begin{table}[h]
\vspace{-2mm}
\scalebox{0.72}{
\begin{tabular}{|c|cc|ccccc|}
\hline 
Method & Rel$\downarrow$ & RMSE$\downarrow$ & 1.05$\uparrow$ & 1.10$\uparrow$ & 1.25$\uparrow$ & $1.25^2$$\uparrow$ & $1.25^3$$\uparrow$ \tabularnewline
\hline 
Smooth & 0.151 & 0.187 & 32.80 & 42.71 & 57.61 & 72.29 & 80.15\tabularnewline
Bilateral \cite{silberman2012indoor} & 0.118 & 0.152 & 34.39 & 46.50 & 61.92 & 75.26 & 81.84 \tabularnewline
Fast \cite{barron2016fast} & 0.127 & 0.154 & 33.65 & 45.08 & 60.36 & 74.52 & 81.79 \tabularnewline
TGV \cite{ferstl2013image} & 0.103 & 0.146 & 37.40 & 48.75 & 62.97 & 75.00 & 81.71 \tabularnewline
Ours & \textbf{0.089} & \textbf{0.116} & \textbf{40.63} & \textbf{51.21} & \textbf{65.35} & \textbf{76.74} & \textbf{82.98} \tabularnewline
\hline
\end{tabular}
}
\vspace{1mm}
\caption{{\bf Comparison to baseline inpainting methods.} Our method significantly outperforms baseline inpainting methods. }
\label{tab:inpainting_comparison}
\vspace{-2mm}
\end{table}


\begin{figure}
\vspace{2mm}
\centering
\includegraphics[width=\linewidth]{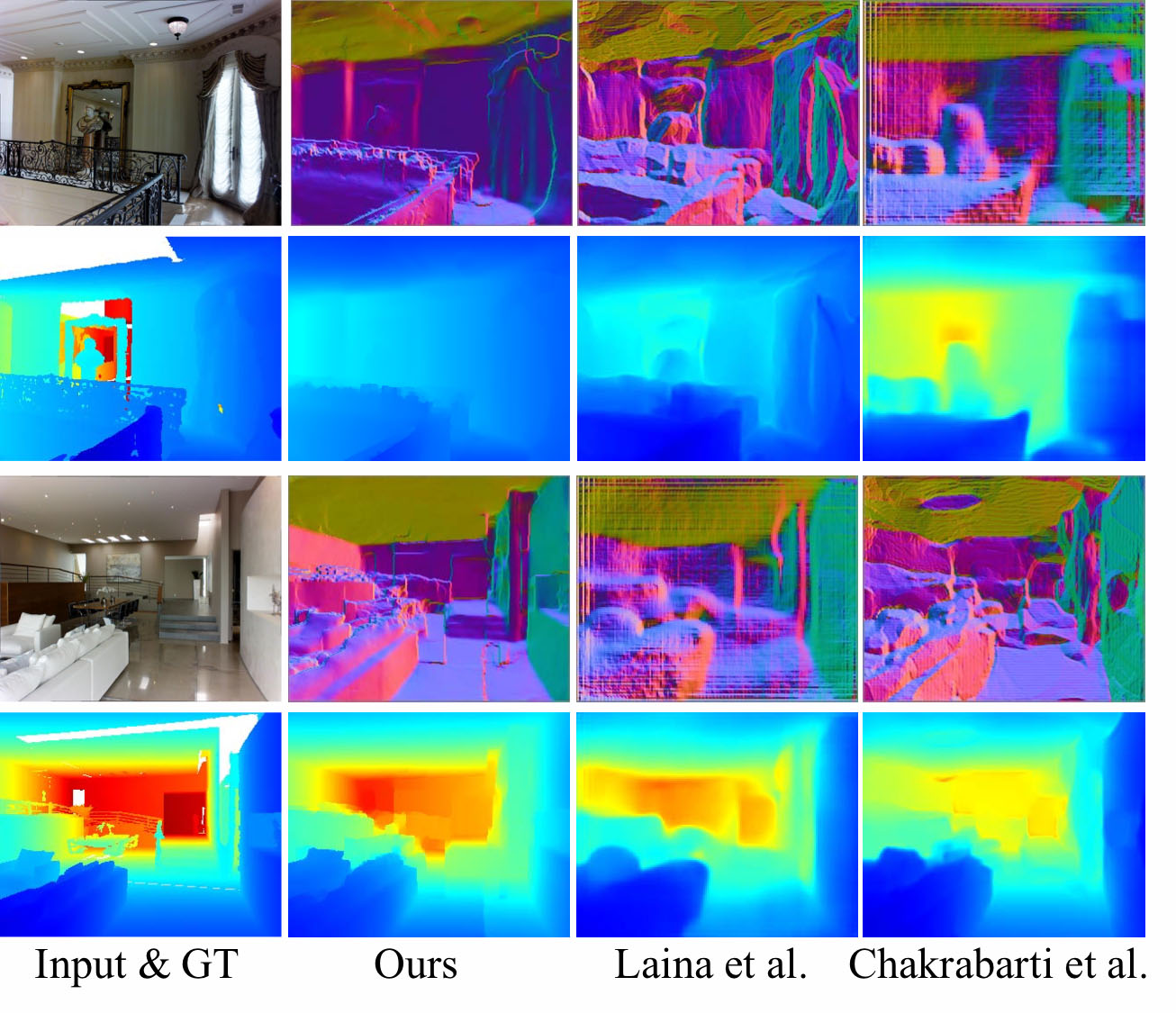}
\caption{{\bf Comparison to deep depth estimation methods.} We compare with the state of the art methods under the depth estimation setting. Our method produces not only accurate depth value but also large scale geometry as reflected in the surface normal.}
\label{fig:estimation_comparison}
\end{figure}

\noindent{\bf Comparison to Depth Estimation Methods}
Table \ref{tab:estimation_comparison} shows results for a study comparing our proposed method to 
previous methods that estimate depth only from color.   We consider comparisons to Chakrabarti et al. 
\cite{chakrabarti2016depth}, whose approach is most similar to ours (it uses predicted derivatives), 
and to Laina et al. \cite{laina2016deeper}, who recently reported state-of-the-art results in experiments 
with NYUv2 \cite{silberman2012indoor}.  We finetune \cite{chakrabarti2016depth} on our dataset, but use pretrained model on NYUv2 for \cite{laina2016deeper} as their training code is not provided.

Of course, these depth estimation methods solve a different problem than
ours (no input depth), and alternative methods have different sensitivities to the scale of depth values,
and so we make our best attempt to adapt both their and our methods to the same setting for fair comparison. 
To do that, we run all methods with only color images as input and then uniformly scale their depth image outputs to align perfectly with the true depth at one random pixel (selected the same for all methods).   In our case, since Equation \ref{eq:objective} is under-constrained without any depth data, we arbitrarily set the middle pixel to a depth of 3 meters during our optimization and then later apply the same scaling as the other methods.
This method focuses the comparison on predicting the ``shape'' of the computed depth image
rather than its global scale. 

Results of the comparison are shown in Figure \ref{fig:estimation_comparison} and Table \ref{tab:estimation_comparison}.
From the qualitative results in Figure \ref{fig:estimation_comparison}, we see that our method reproduces both the structure of the scene and the fine details best -- even when given only one pixel of raw depth.
According to the quantitative results shown in Table \ref{tab:estimation_comparison}, our method is 23-40\% better than the others, regardless of whether evaluation pixels have observed depth (Y)
or not (N).  These results suggest that predicting
surface normals is a promising approach to depth estimation as well.

\begin{table}
\vspace{-2mm}
\scalebox{0.72}{
\begin{tabular}{|c|c|cc|ccccc|}
\hline 
Obs & Meth & Rel$\downarrow$ & RMSE$\downarrow$ & 1.05$\uparrow$ & 1.10$\uparrow$ & 1.25$\uparrow$ & $1.25^2$$\uparrow$ & $1.25^3$$\uparrow$\tabularnewline
\hline 
\multirow{3}{*}{Y} 
 & \cite{laina2016deeper} & 0.190 & 0.374 & 17.90 & 31.03 & 54.80 & 75.97 & 85.69 \tabularnewline
 & \cite{chakrabarti2016depth} & 0.161 & 0.320 & 21.52 & 35.5 & 58.75 & 77.48 & 85.65 \tabularnewline
 & Ours & \textbf{0.130} & \textbf{0.274} & \textbf{30.60} & \textbf{43.65} & \textbf{61.14} & \textbf{75.69} & \textbf{82.65} \tabularnewline
\hline
\multirow{3}{*}{N} 
 & \cite{laina2016deeper} & 0.384 & 0.572 & 8.86 & 16.67 & 34.64 & 55.60 & 69.21 \tabularnewline
 & \cite{chakrabarti2016depth} & 0.352 & 0.610 & 11.16 & 20.50 & 37.73 & 57.77 & 70.10 \tabularnewline
 & Ours & \textbf{0.283} & \textbf{0.537} & \textbf{17.27} & \textbf{27.42} & \textbf{44.19} & \textbf{61.80} & \textbf{70.90} \tabularnewline  
\hline 
\end{tabular}
}
\vspace{1mm}
\caption{{\bf Comparison to deep depth estimation methods.} We compare with Laina et al. \cite{laina2016deeper} and Chakrabarti et al.\cite{chakrabarti2016depth}. All the methods perform worse on unobserved pixels than the observed pixels, which indicates unobserved pixels are harder. Our method significantly outperform other methods.}
\label{tab:estimation_comparison}
\end{table}

\vspace{-1mm}
\section{Conclusion}
\label{sec:conclusion}
\vspace{-1mm}
This paper describes a deep learning framework for completing the depth channel of an RGB-D image acquired with a commodity RGB-D camera. It provides two main research contributions. First, it proposes to complete depth with a two stage process where surface normals and occlusion boundaries are predicted from color, and then completed depths are solved from those predictions. Second, it learns to complete depth images by supervised training on data rendered from large-scale surface reconstructions. During tests with a new benchmark, we find the proposed approach outperforms previous baseline approaches for depth inpainting and estimation.

\clearpage

{\small
\bibliographystyle{ieee}
\bibliography{egbib}
}

\clearpage
\appendixpage
\begin{appendices}

This document contains further implementation details and results of ablation studies, cross-dataset experiments, and comparisons to other inpainting methods that would not fit in the main paper.


\section{Further Implementation Details}

This section provides extra implementation details for our methods.
All data and code will be released upon the acceptance to ensure reproducibility. 

\subsection{Mesh reconstruction and rendering}

For every scene in the Matterport3D dataset, meshes were reconstructed and rendered to provide ``completed depth images'' using the following process.   First, each house was manually partitioned into regions roughly corresponding to rooms using an interactive floorplan drawing interface.   Second, a dense point cloud was extracted containing RGB-D points (pixels) within each region, excluding pixels whose depth is beyond 4 meters from the camera (to avoid noise in the reconstructed mesh).  Third, a mesh was reconstructed from the points of each region using Screened Poisson Surface Reconstruction \cite{kazhdan2013screened} with octree depth 11.  The meshes for all regions were then merged to form the final reconstructed mesh $M$ for each scene.  ``Completed depth images'' were then created for each of the original RGB-D camera views by rendering $M$ from that view using OpenGL and reading back the depth buffer.  

\begin{figure}[t]
\centering
\includegraphics[width=0.49\linewidth]{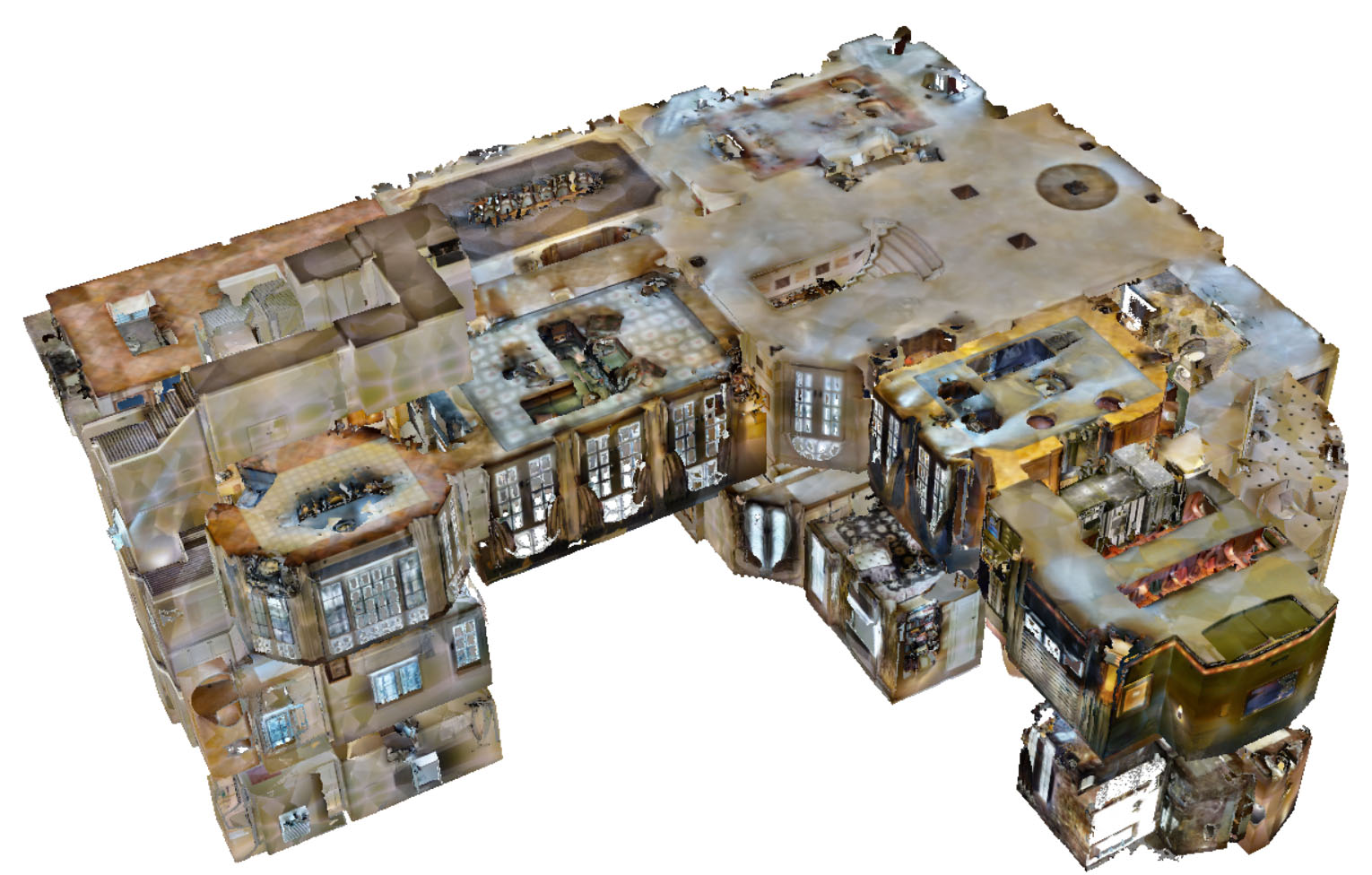}
\includegraphics[width=0.49\linewidth]{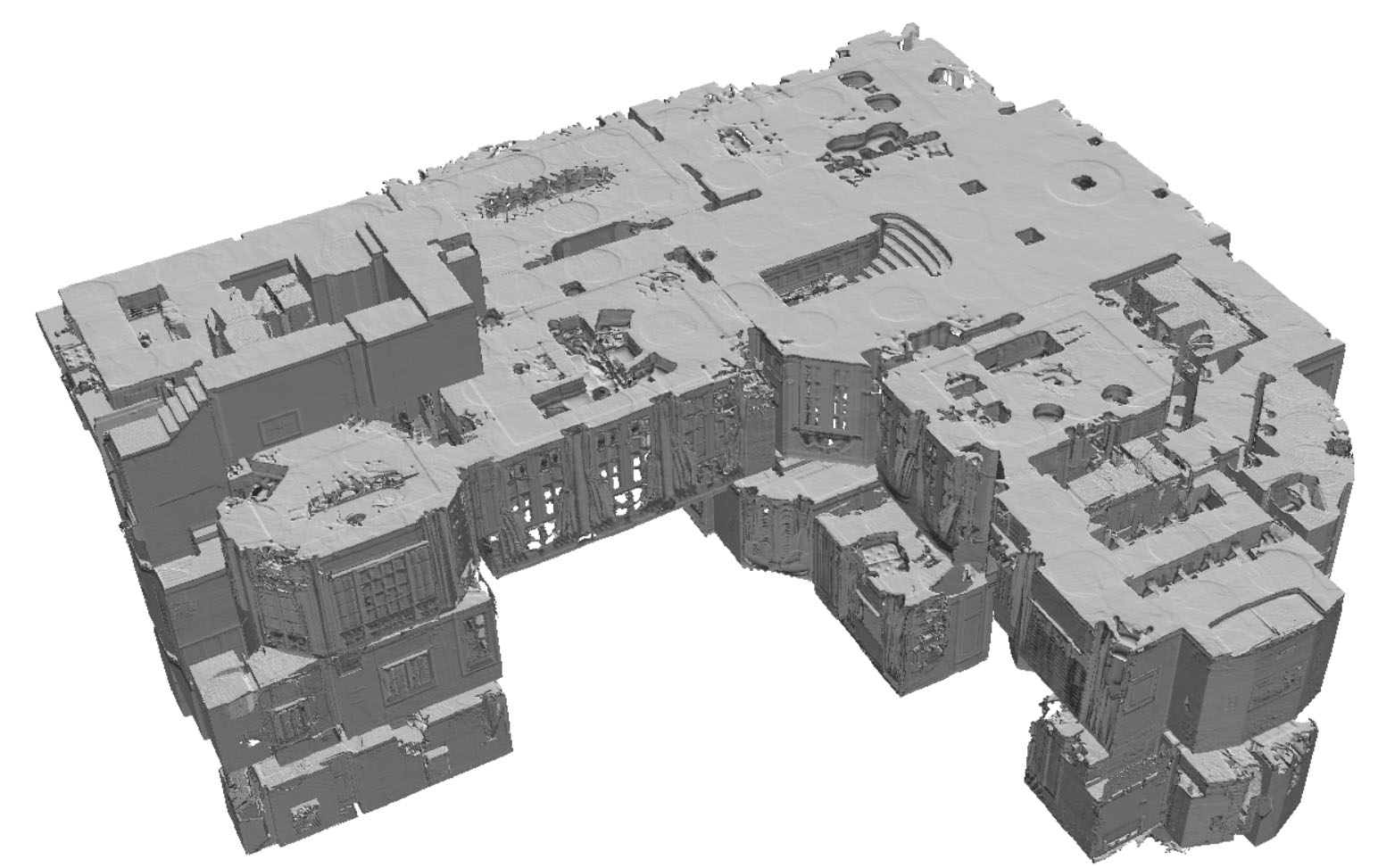} \\
\includegraphics[width=0.49\linewidth]{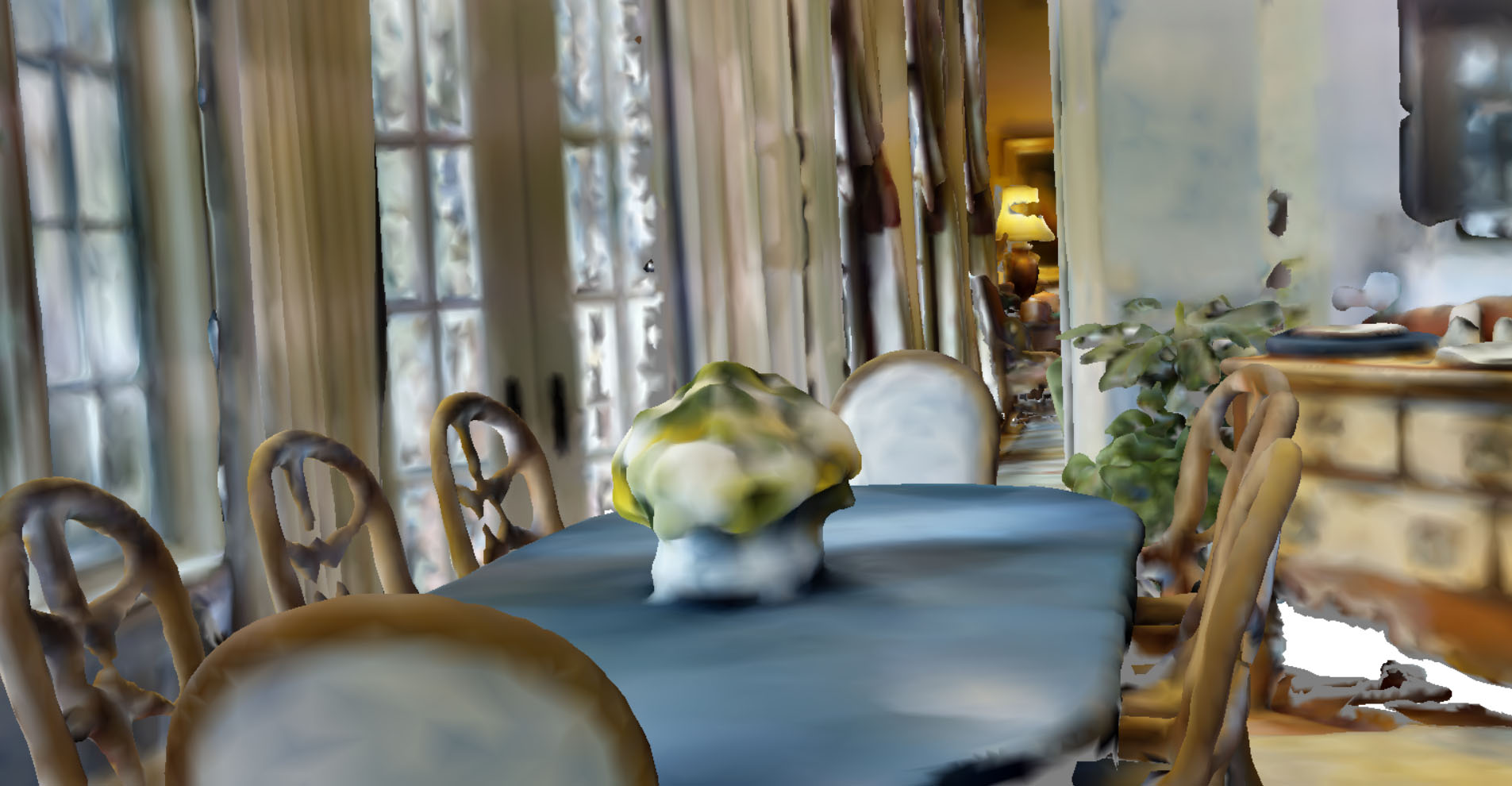}
\includegraphics[width=0.49\linewidth]{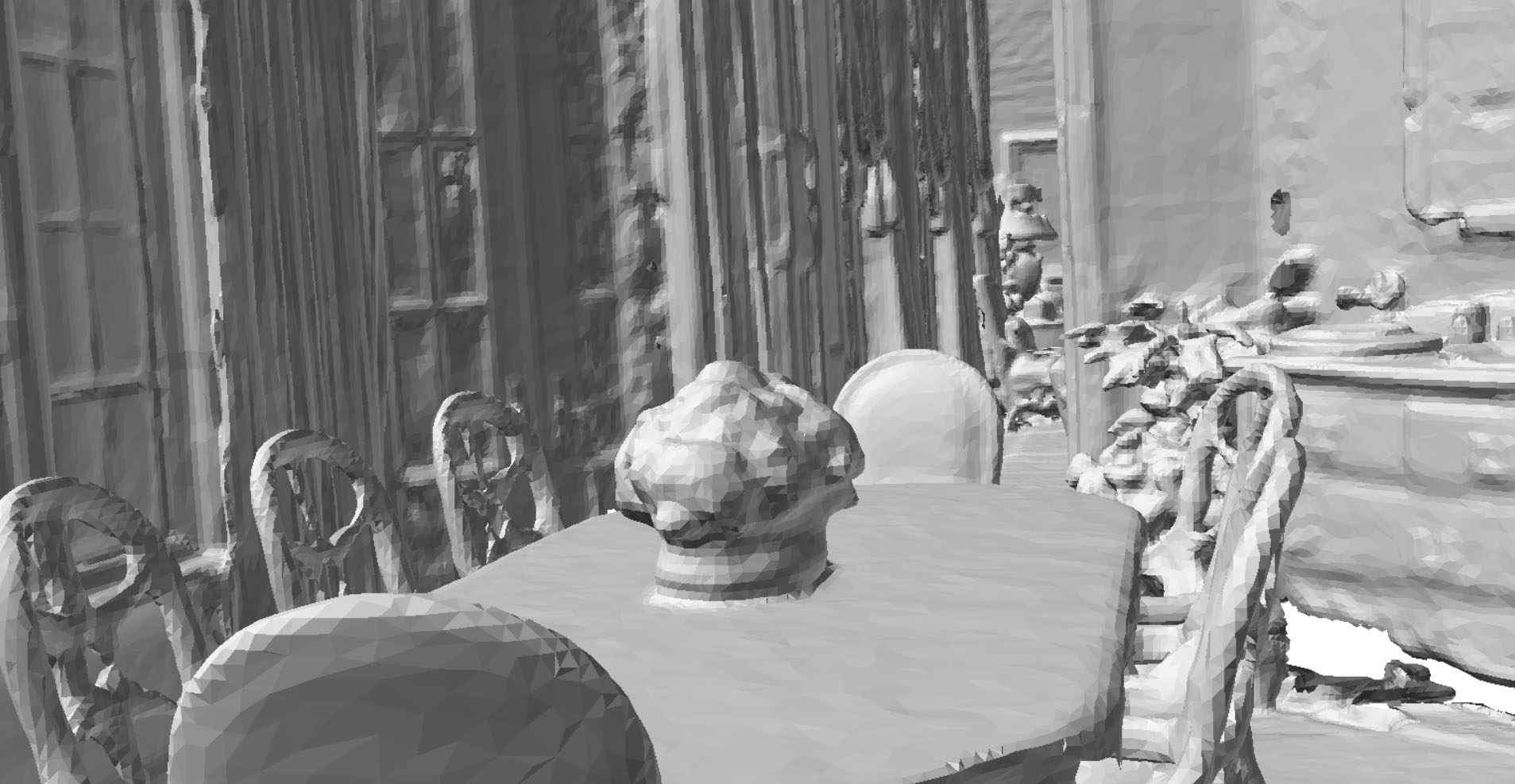}
\caption{{\bf Reconstructed mesh for one scene.} The mesh used to
render completed depth images is shown from an outside view (top) and
inside view (bottom), rendered with vertex colors (left) and flat shading (right).}
\label{fig:mesh}
\end{figure}

Figure \ref{fig:mesh} shows images of a mesh produced with this process.   The top row shows exterior views covering the entire house (vertex colors on the left, flat shading on the right).  The bottom row shows a close-up image of the mesh from an interior view.   Though the mesh is not perfect, it has 12.2M triangles reproducing most surface details.   Please note that the mesh is complete where holes typically occur in RGB-D images (windows, shiny table tops, thin structures of chairs, glossy surfaces of cabinet, etc.).   Please also note the high level of detail for surfaces distant to the camera (e.g., furniture in the next room visible through the doorway).

\subsection{Network architecture}
All the networks used for this project are derived from the surface normal estimation model proposed in Zhang et.al \cite{zhang2017physically} with the following modifications.

\noindent \paragraph{Input}
Depending on what is the input, the network takes data with different channels at the first convolution layer.
\begin{itemize}
    \item Color. The color is a 3-channel tensor with R,G,B for each. The intensity values are normalized to [-0.5 0.5]. We use a bi-linear interpolation to resize color image if necessary.
    \item Depth. The absolute values of depth in meter are used as input. The pixels with no depth signal from sensor are assigned a value of zero. To resolve the ambiguity between ``missing'' and ``0 meter'', a binary mask indicating the pixels that have depth from sensor is added as an additional channel as suggested in Zhang et.al \cite{zhang2017real}. Overall, the depth input contains 2 channels (absolute depth and binary valid mask) in total. To prevent inaccurate smoothing, we use the nearest neighbor search to resize depth image. 
    \item Color+Depth. The input in this case is the concatenation of the color and depth as introduced above. This results in a 5-channel tensor as the input.
\end{itemize}

\noindent \paragraph{Output}
The network for absolute depth, surface normal, and depth derivative outputs results with 1, 3, and 8 channels respectively.
The occlusion boundary detection network generates 3 channel outputs representing the probability of each pixel belonging to ``no edge'', ``depth crease'', and ``occlusion boundary''.

\noindent \paragraph{Loss}
Depth, surface normal, and derivative are predicted as regression tasks.
The SmoothL1 loss\footnote{https://github.com/torch/nn/blob/master/doc/criterion.md\#nn.Smoot-\\hL1Criterion} is used for training depth and derivative, and the cosine embedding loss\footnote{https://github.com/torch/nn/blob/master/doc/criterion.md\#nn.Cosine-\\EmbeddingCriterion} is used for training surface normal.
The occlusion boundary detection is formulated into a classification task, and cross entropy loss\footnote{https://github.com/torch/nn/blob/master/doc/criterion.md\#nn.CrossE-\\ntropyCriterion} is used.
The last two batch normalization layers are removed because this results in better performance in practice.

\subsection{Training schema}

The neural network training and testing are implemented in Torch.
For all the training tasks, RMSprop optimization algorithm is used.
The momentum is set to 0.9, and the batch size is 1.
The learning rate is set to 0.001 initially and reduce to half every 100K iterations.
All the models converge within 300K iterations.

\section{Further Experimental Results}

This section provides extra experimental results, including 
ablation studies, cross-dataset experiments, and comparisons to
other depth completion methods.

\subsection{Ablation Studies}

Section 4.1 of the paper provides results of ablation studies aimed at investigating how different
test inputs, training data, loss functions, depth representations, and optimization methods affect our depth prediction results.   This section provides further results of that type.

\begin{figure*}[ht]
\centering
\includegraphics[width=\linewidth]{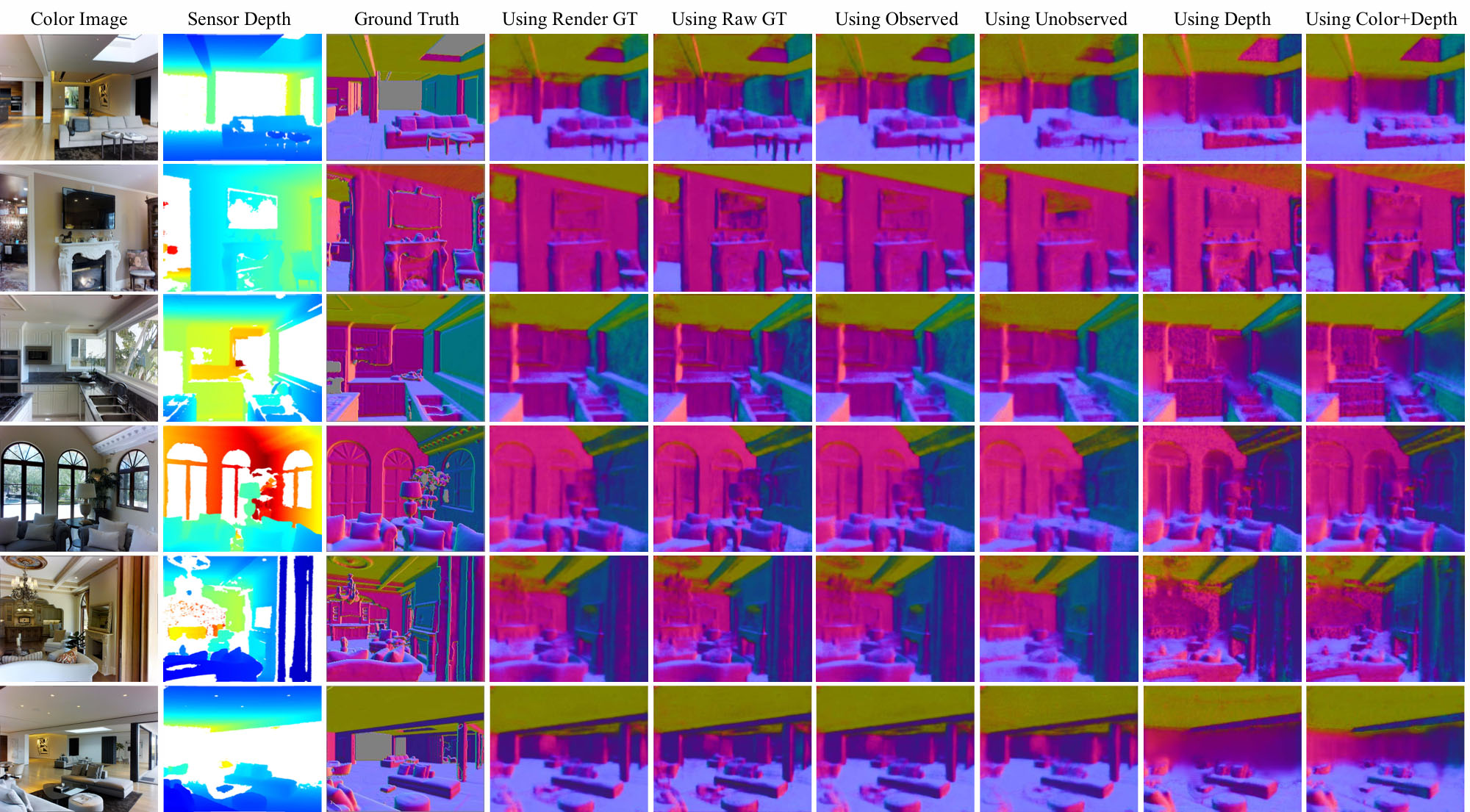}
\caption{{\bf Comparison of normal estimation with different training settings.} The 4-th column shows the output of the model trained using only color as input and the rendered depth from all pixels as supervision, which is the setting we chose for our system. Comparatively, it generates better surface normal than other alternative training settings. } 
\label{fig:ablation_normal}
\end{figure*}

More qualitative results about surface normal estimation model trained from different setting are shown in Figure \ref{fig:ablation_normal}.
Comparatively, training the surface normal estimation model with our setting (i.e. using only color image as input, all available pixels with rendered depth as supervision, the 4-th column in the figure) achieves the best quality of prediction, and hence benefits the global optimization for depth completion.


\begin{table}[t]
\scalebox{0.72}{
\begin{tabular}{|c|c|cc|ccccc|}
\hline 
Input & Rep & Rel$\downarrow$ & RMSE$\downarrow$ & 1.05$\uparrow$ & 1.10$\uparrow$ & 1.25$\uparrow$ & $1.25^2$$\uparrow$ & $1.25^3$$\uparrow$ \tabularnewline
\hline 
C & D & 0.408 & 0.500 & 6.49 & 12.80 & 30.01 & 54.44 & 72.88 \tabularnewline
C & 1/D & 0.412 & 0.492 & 6.86 & 12.88 & 28.99 & 54.51 & 73.13 \tabularnewline
\hline
D & D & 0.167 & 0.241 & 16.43 & 31.13 & 57.62 & 75.63 & \textbf{84.01} \tabularnewline 
D & 1/D & 0.199 & 0.255 & 14.06 & 27.32 & 53.70 & 74.19 & 83.85 \tabularnewline
\hline
\multicolumn{2}{|c|}{Ours} & \textbf{0.089} & \textbf{0.116} & \textbf{40.63} & \textbf{51.21} & \textbf{65.35} & \textbf{76.74} & 82.98\tabularnewline
\hline
\end{tabular}
}
\vspace{1mm}
\caption{{\bf Comparison of different depth representations.} Predicting either depth (D) or disparity (1/D) provides worse results than predicting surface normals and solving for depth (Ours) for either color or depth inputs.}
\label{tab:disparity}
\end{table}


\begin{table*}[t]
\centering
\scalebox{0.765}{
\begin{tabular}{|c|ccc|cc|ccccc|cc|ccc|}
\hline
\multirow{2}{*}{Comparison} & \multicolumn{3}{|c|}{Setting} & \multicolumn{7}{|c|}{Depth Completion} & \multicolumn{5}{|c|}{Surface Normal Estimation} \tabularnewline
\cline{2-16}
 & Input & Target & Pixel & Rel$\downarrow$ & RMSE$\downarrow$ & 1.05$\uparrow$ & 1.10$\uparrow$ & 1.25$\uparrow$ & $1.25^2$ $\uparrow$ & $1.25^3$ $\uparrow$ & Mean$\downarrow$ & Median$\downarrow$ & 11.25$\uparrow$ & 22.5$\uparrow$ & 30$\uparrow$ \tabularnewline
\hline 
\multirow{2}{*}{Target} & Color & \textit{Raw} & Both & 0.094 & 0.123 & 39.84 & 50.40 & 64.68 & 76.38 & 82.80 & 32.87 & 18.70 & 34.2 & 55.7 & 64.3 \tabularnewline
  & Color & \textit{Render} & Both & 0.089 & 0.116 & 40.63 & 51.21 & 65.35 & 76.64 & 82.98 & 31.13 & 17.28 & 37.7 & 58.3 & 67.1 \tabularnewline
\hline 
\multirow{3}{*}{Pixel} & Color & Render & \textit{Observed}    & 0.091 & 0.121 & 40.31 & 50.88 & 64.92 & 76.50 & 82.91 & 32.16 & 18.44 & 34.7 & 56.4 & 65.5 \tabularnewline
  & Color & Render & \textit{Unobserved}  & 0.090 & 0.119 & 40.71 & 51.22 & 65.21 & 76.59 & 83.04 & 31.52 & 17.70 & 35.4 & 57.7 & 66.6  \tabularnewline
  & Color & Render & \textit{Both}        & 0.089 & 0.116 & 40.63 & 51.21 & 65.35 & 76.64 & 82.98 & 31.13 & 17.28 & 37.7 & 58.3 & 67.1 \tabularnewline
\hline
\multirow{3}{*}{Input} & \textit{Depth} & Render & Both   &  0.107 & 0.165 & 38.89 & 48.54 & 61.12 & 73.57 & 80.98 & 35.08 & 23.07 & 27.6 & 49.1 & 58.6 \tabularnewline
  & \textit{Both} & Render & Both  & 0.090 & 0.124 & 40.13 & 51.26 & 64.84 & 76.46 & 83.05 & 35.30 & 23.59 & 26.7 & 48.5 & 58.1 \tabularnewline
  & \textit{Color} & Render & Both       & 0.089 & 0.116 & 40.63 & 51.21 & 65.35 & 76.64 & 82.98 & 31.13 & 17.28 & 37.7 & 58.3 & 67.1 \tabularnewline
\hline
\end{tabular}
}
\vspace{1mm}
\caption{Ablation studies.  Evaluations of estimated surface normals and solved depths using different training inputs and losses. For the sake of comparison, Table 1 from main paper is copied in the last three rows as comparison across different inputs. }
\label{tab:ablation}
\end{table*}


\vspace{1mm} \noindent \paragraph{What kind of ground truth is better?}
This test studies what normals should be used as supervision for the loss when training the surface prediction network.   We experimented with normals computed from raw depth images and with normals computed from the rendered mesh.  The result in the top two rows of Table \ref{tab:ablation} (Comparison:Target) shows that the model trained on rendered depth performs better the the one from raw depth.  The improvement seems to come partly from having training pixels for unobserved regions and partly from more accurate depths (less noise).  

\vspace{1mm} \noindent \paragraph{What loss should be used to train the network?}
This test studies which pixels should be included in the loss when training the surface prediction network.  We experimented with using only the unobserved pixels, using only the observed pixels, and both as supervision.  The three models were trained separately in the training split of our new dataset and then evaluated versus the rendered normals in the test set.   The quantitative results in the last three rows of Table \ref{tab:ablation} (Comparison:Pixel) show that models trained with supervision from both observed and unobserved pixels (bottom row) works slightly better than the one trained with only the observed pixels or only the unobserved pixels.   This shows that the unobserved pixels indeed provide additional information.

\vspace{1mm} \noindent \paragraph{What kind of depth representation is best?}
Several depth representations were considered in the paper (normals, derivatives, depths, etc.).   This section provides further results regarding direct prediction of depth and disparity (i.e. one over depth) to augment/fix results in Table 2 of the paper. 

Actually, the top row of Table 2 of the paper (where the Rep in column 2 is `D') is mischaracterized as direct prediction of depth from color -- it is actually direct prediction of complete depth from input depth.  That was a mistake.   Sorry for the confusion.   The correct result is in the top line of Table \ref{tab:disparity} of this document (Input=C, Rep=D).   The result is quite similar and does not change any conclusions: predicting surface normals and then solving for depth is better than predicting depth directly (Rel = 0.089 vs. 0.408).

We also consider prediction of disparity rather than depth, as suggested in Chakrabarti et.al and other papers \cite{chakrabarti2016depth}.  We train models to estimate disparity directly from color and raw depth respectively.
The results can be seen in Table \ref{tab:disparity}.  We find that estimating disparity results in performance that is not better than estimating depth when given either color or depth as input for our depth completion application.


\subsection{Cross-Dataset Experiments}

This test investigates whether it is possible to train our method on one dataset and then use it effectively for another.   

\vspace{1mm} \noindent \paragraph{Matterport3D and ScanNet}
We first conduct experiments between Matterport3D and ScanNet datasets.   Both have 3D surface reconstructions for large sets of environments ($\sim$1000 rooms each) and thus provide suitable training data for training and test our method with rendered meshes.    We train a surface normal estimation model separately on each dataset, and then use it without fine tuning to perform depth completion for the test set of the other.
The quantitative results are shown in Table \ref{tab:cross_dataset}.
As expected, the models work best on the test dataset matching the source of the training data.  Actually, the model trained from Matterport3D has a better generalization capability compared to the model trained from ScanNet, which is presumably because the Matterport3D dataset has a more diverse range of camera viewpoints.   However, interestingly, both models work still reasonably well when run on the other dataset, even though they were not fine-tuned at all.  We conjecture this is because our surface normal prediction model is trained only on color inputs, which are relatively similar between the two datasets.  Alternative methods using depth as input would probably not generalize as well due to the significant differences between the depth images of the two datasets.

\begin{table*}[hbt]
\centering
\scalebox{1}{
\begin{tabular}{|c|c|cc|ccccc|}
\hline 
Train & Test & Rel & RMSE & 1.05 & 1.10 & 1.25 & $1.25^2$ & $1.25^3$ \tabularnewline
\hline 
Matterport3D & Matterport3D & 0.089 & 0.116 & 40.63 & 51.21 & 65.35 & 76.74 & 82.98 \tabularnewline
ScanNet & Matterport3D & 0.098 & 0.128 & 37.96 & 49.79 & 64.01 & 76.04 & 82.64 \tabularnewline
\hline 
Matterport3D & Scannet & 0.042 & 0.065 & 52.91 & 65.83 & 81.20 & 90.99 & 94.94\tabularnewline
ScanNet & ScanNet & 0.041 & 0.064 & 53.33 & 66.02 & 81.14 & 90.92 & 94.92 \tabularnewline
\hline 
\end{tabular}
}
\vspace{1mm}
\caption{{\bf Cross-dataset performance.} We trained surface normal estimation models on each dataset, Matterport3D and ScanNet, respectively and test on both. Models work the best on the dataset where it is trained from. Model trained from Matterport3D shows better generalization capability than the one from ScanNet.}
\label{tab:cross_dataset}
\end{table*}

\vspace{1mm} \noindent \paragraph{Intel RealSense Depth Sensor}

\begin{figure*}[ht]
\centering
\vspace{-6mm}
\centering
\includegraphics[width=\linewidth]{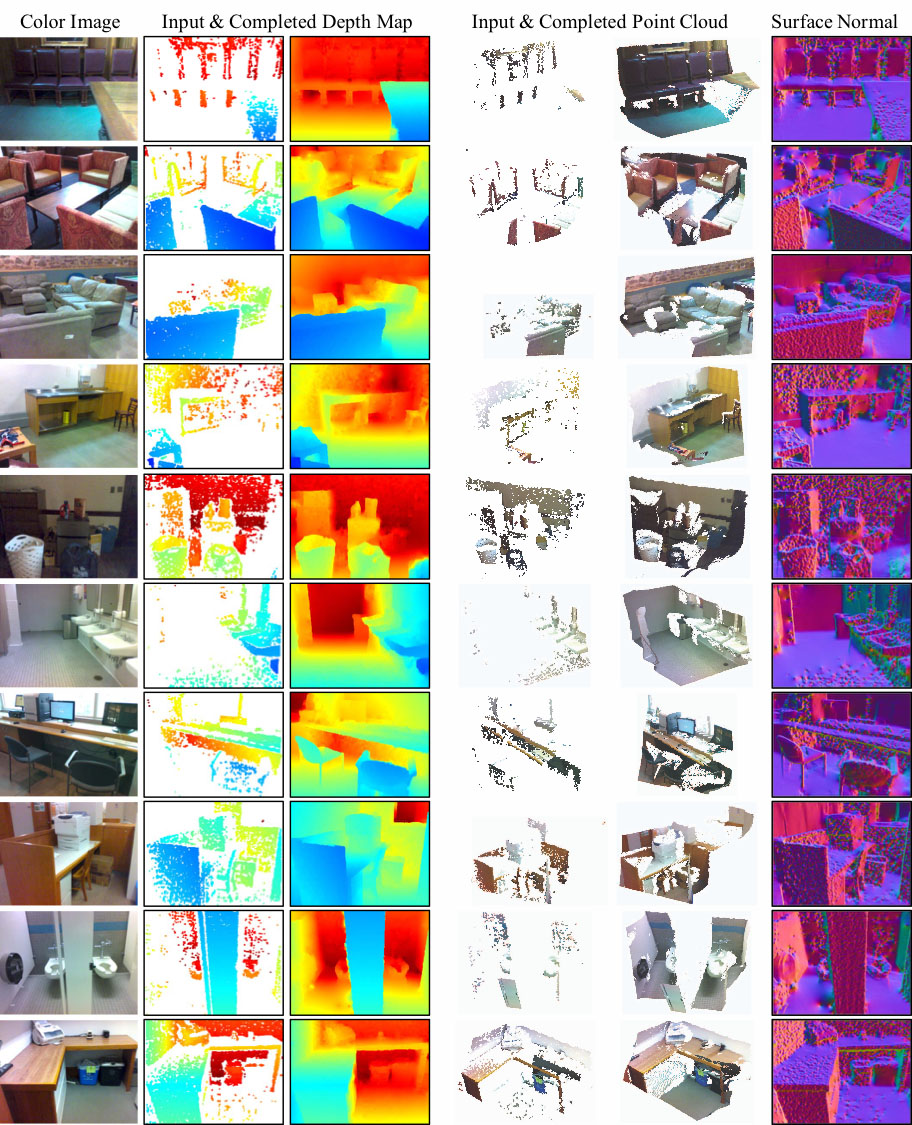}
\caption{{\bf Our results on RealSense data.} We run a model trained from ScanNet dataset and test on RGBD images captured by Intel RealSense without finetune. From left to right, we show the input color image, input depth image, completed depth image using our method, the point cloud visualization of the input and completed depth map, and the surface normal converted from the completed depth. Our method generates good results for depth completion.} 
\label{fig:realsense}
\end{figure*}

The depth map from Intel RealSense has better quality in short range but contains more missing area compared to Structure Sensor \cite{structio} and Kinect \cite{kinect}.
The depth signal can be totally lost or extremely sparse for distant area and surface with special materials, e.g. shinny, dark.
We train a surface normal estimation model from ScanNet dataset \cite{dai2017scannet} and directly evaluate on the RGBD images captured by Intel RealSense from SUN-RGBD dataset \cite{song2015sun} without any finetuning.
The results are shown in Figure \ref{fig:realsense}.
From left to right, we show the input color image, input depth image, completed depth image using our method, the point cloud visualization of the input and completed depth map, and the surface normal converted from the completed depth.
As can be seen, the depth from RealSense contains more missing area than Matterport3D and ScanNet, yet our model still generates decent results.
This again shows that our method can effectively run on RGBD images captured from various of depth sensors with significantly different depth patterns.

\subsection{Comparisons to Depth Inpainting Methods}

Section 4.2 of the paper provides comparisons to alternative methods for depth inpainting.   This section provides further results of that type in Table \ref{tab:inpainting_comparison}. In this additional study, we compare with the following methods:

\begin{itemize}
    \item {\bf DCT \cite{garcia2010robust}:} fill in missing values by solving the penalized least squares of a linear system using discrete cosine transform using the code from Matlab Central \footnote{https://www.mathworks.com/matlabcentral/fileexchange/27994-inpaint-over-missing-data-in-1-d--2-d--3-d--nd-arrays}.
    \item {\bf FCN \cite{mao2016image}:} train an FCN with symmetric shortcut connection to take raw depth as input and generate completed depth as the output using the code from Zhang et.al \cite{zhang2017physically}.
    \item {\bf CE \cite{pathak2016context}:} train the context encoder of Pathak et.al to inpaint depth images using the code from Github \footnote{https://github.com/pathak22/context-encoder}.
\end{itemize}

The results of DCT \cite{garcia2010robust} are similar to other inpainting comparisons provided in the paper.  They mostly interpolate holes.

The results of FCN and CE show that methods designed for inpainting color are not very effective at inpainting depth.   As already described in the paper, methods that learn depth from depth using an FCN can be lazy and only learn to reproduce and interpolate provided depth.   However, the problems are more subtle than that, as depth data has many characteristics different from color.  For starters, the context encoder has a more shallow generator and lower resolution than our network, and thus generates blurrier depth images than ours.  More significantly, the fact that ground-truth depth data can have missing values complicates the training of the discriminator network in the context encoder (CE) -- in a naive implementation, the generator would be trained to predict missing values in order to fool the discriminator.  We tried multiple approaches to circumvent this problem, including propagating gradients on only unobserved pixels, filling a mean depth value in the missing area.   We find that none of them work as well as our method. 

\begin{figure*}[ht]
\vspace{-6mm}
\centering
\includegraphics[width=\linewidth]{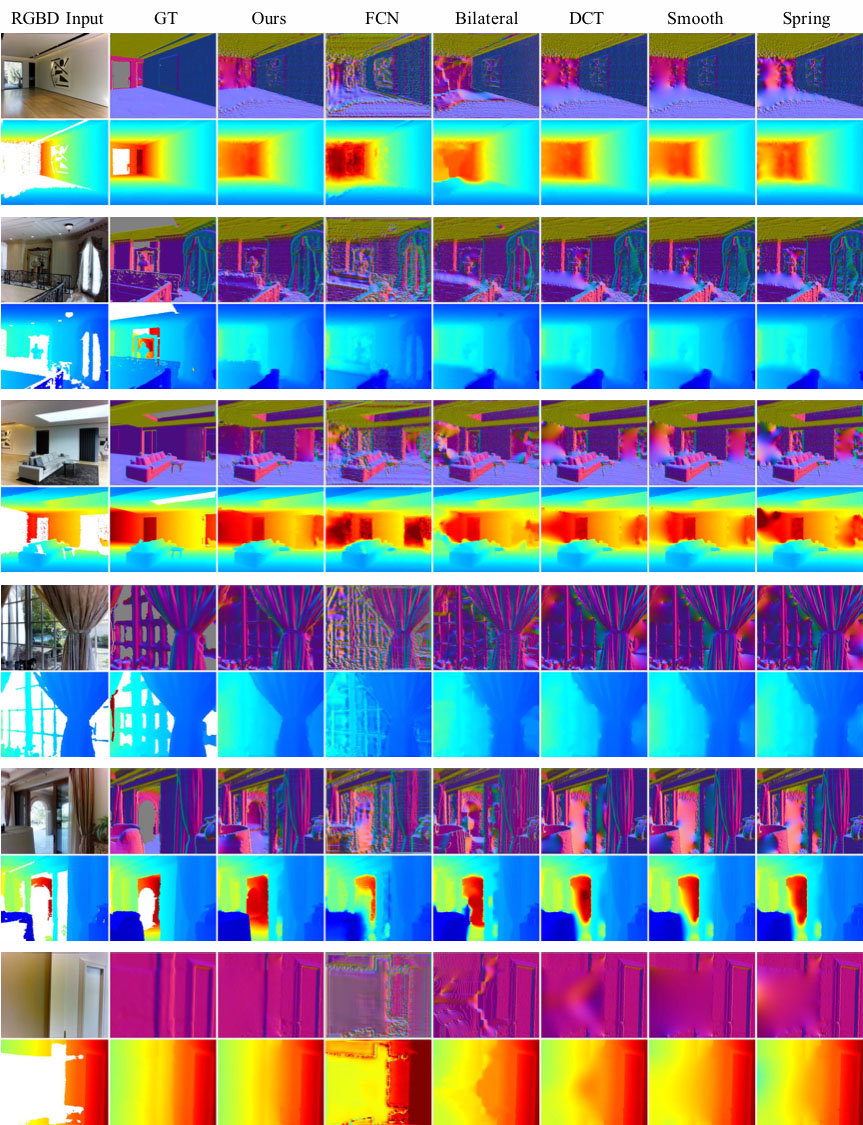}
\caption{{\bf More results and comparison to inpainting methods.} Each example is shown in two rows, where the second row shows the input, ground truth, and completed depth, whereas the first row shows the surface normal of each corresponding depth map on the second row to highlight details. Our method in general works better than other inpainting methods.} 
\label{fig:inpainting_comparsion1}
\end{figure*}

\begin{figure*}[ht]
\vspace{-6mm}
\centering
\includegraphics[width=\linewidth]{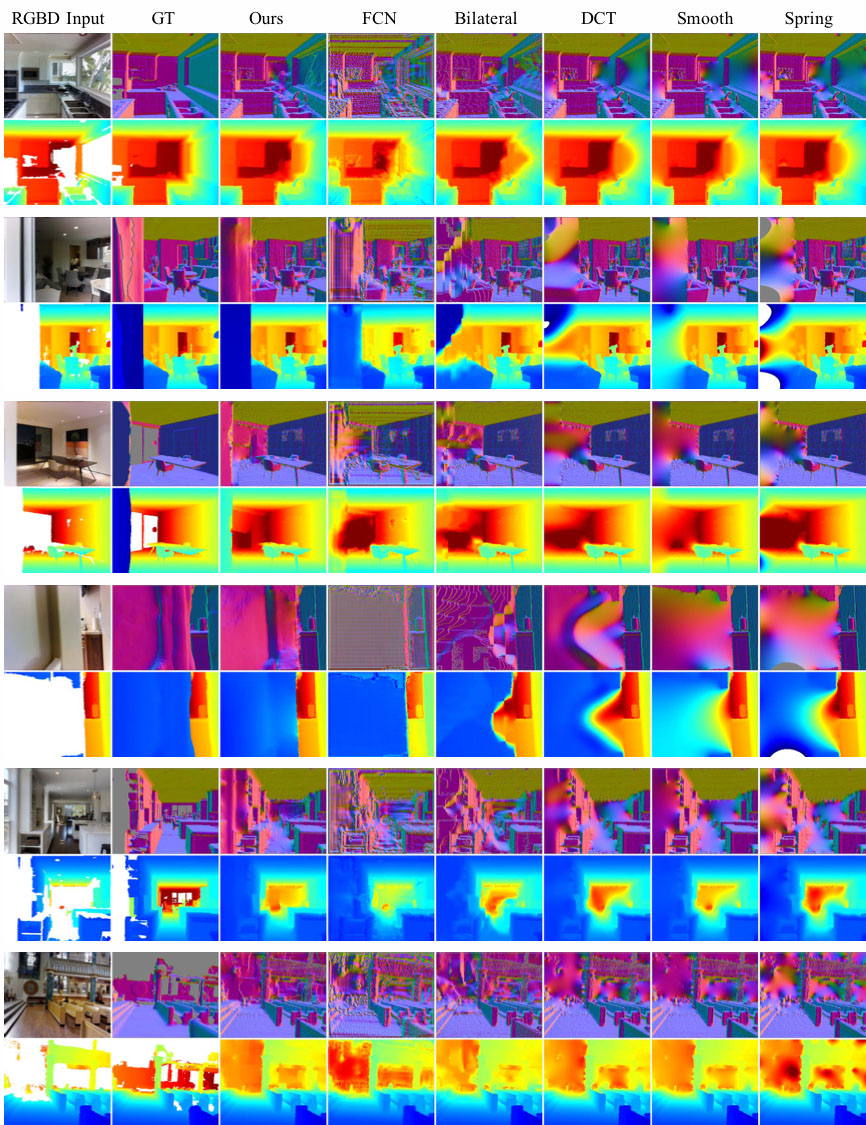}
\caption{{\bf More results and comparison to inpainting methods.} Each example is shown in two rows, where the second row shows the input, ground truth, and completed depth, whereas the first row shows the surface normal of each corresponding depth map on the second row to highlight details. Our method in general works better than other inpainting methods.} 
\label{fig:inpainting_comparsion2}
\end{figure*}

\begin{figure*}[ht]
\vspace{-6mm}
\centering
\includegraphics[width=\linewidth]{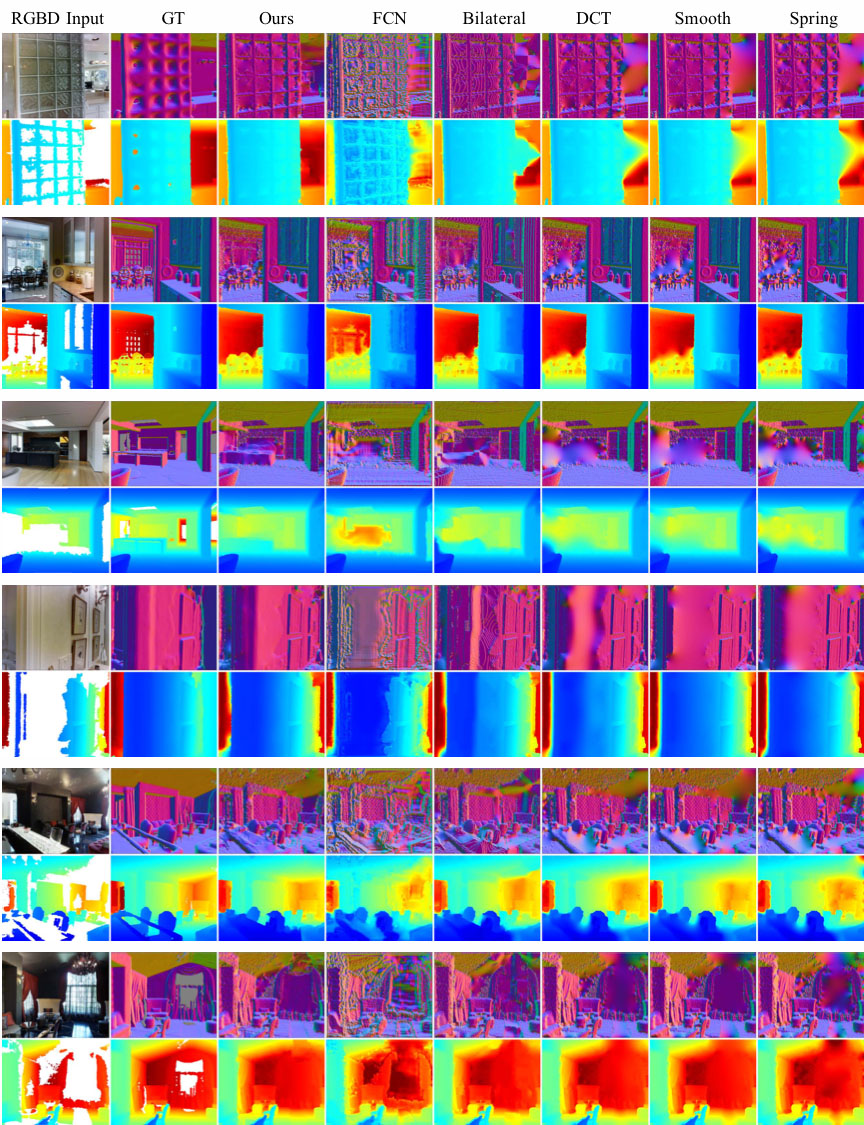}
\caption{{\bf More results and comparison to inpainting methods.} Each example is shown in two rows, where the second row shows the input, ground truth, and completed depth, whereas the first row shows the surface normal of each corresponding depth map on the second row to highlight details. Our method in general works better than other inpainting methods.} 
\label{fig:inpainting_comparsion3}
\end{figure*}

More results of our method and comparison to other inpainting methods can be found in Figure \ref{fig:inpainting_comparsion1},\ref{fig:inpainting_comparsion2},\ref{fig:inpainting_comparsion3} in the end of this paper. Each two rows shows an example, where the 2nd row shows the completed depth of different methods, and 1st row shows their corresponding surface normal for purpose of highlighting details and 3D geometry. For each example, we show the input, ground truth, our result, followed by the results of FCN \cite{mao2016image}, joint bilateral filter \cite{silberman2012indoor}, discrete cosine transform \cite{garcia2010robust}, optimization with only smoothness, and PDE \cite{derrico2017inpaint}.
As can be seen, our method generates better large scale planar geometry and sharper object boundary.

\begin{table}[ht]
\vspace{-2mm}
\scalebox{0.72}{
\begin{tabular}{|c|cc|ccccc|}
\hline 
Method & Rel$\downarrow$ & RMSE$\downarrow$ & 1.05$\uparrow$ & 1.10$\uparrow$ & 1.25$\uparrow$ & $1.25^2$$\uparrow$ & $1.25^3$$\uparrow$ \tabularnewline
\hline 
Smooth & 0.151 & 0.187 & 32.80 & 42.71 & 57.61 & 72.29 & 80.15\tabularnewline
Bilateral \cite{silberman2012indoor} & 0.118 & 0.152 & 34.39 & 46.50 & 61.92 & 75.26 & 81.84 \tabularnewline
Fast \cite{barron2016fast} & 0.127 & 0.154 & 33.65 & 45.08 & 60.36 & 74.52 & 81.79 \tabularnewline
TGV \cite{ferstl2013image} & 0.103 & 0.146 & 37.40 & 48.75 & 62.97 & 75.00 & 81.71 \tabularnewline
\hline
Garcia et.al \cite{garcia2010robust} & 0.115 & 0.144 & 36.78 & 47.13 & 61.48 & 74.89 & 81.67 \tabularnewline
FCN \cite{zhang2017physically} & 0.167 & 0.241 & 16.43 & 31.13 & 57.62 & 75.63 & 84.01 \tabularnewline 
Ours & \textbf{0.089} & \textbf{0.116} & \textbf{40.63} & \textbf{51.21} & \textbf{65.35} & \textbf{76.74} & \textbf{82.98} \tabularnewline
\hline
\end{tabular}
}
\vspace{1mm}
\caption{{\bf Comparison to baseline inpainting methods.} For the sake of comparison, we copy the methods compared in the main paper in the same table. Our method significantly outperforms baseline inpainting methods. }
\label{tab:inpainting_comparison}
\vspace{-2mm}
\end{table}

\begin{figure*}[ht]
\vspace{-6mm}
\centering
\includegraphics[width=\linewidth]{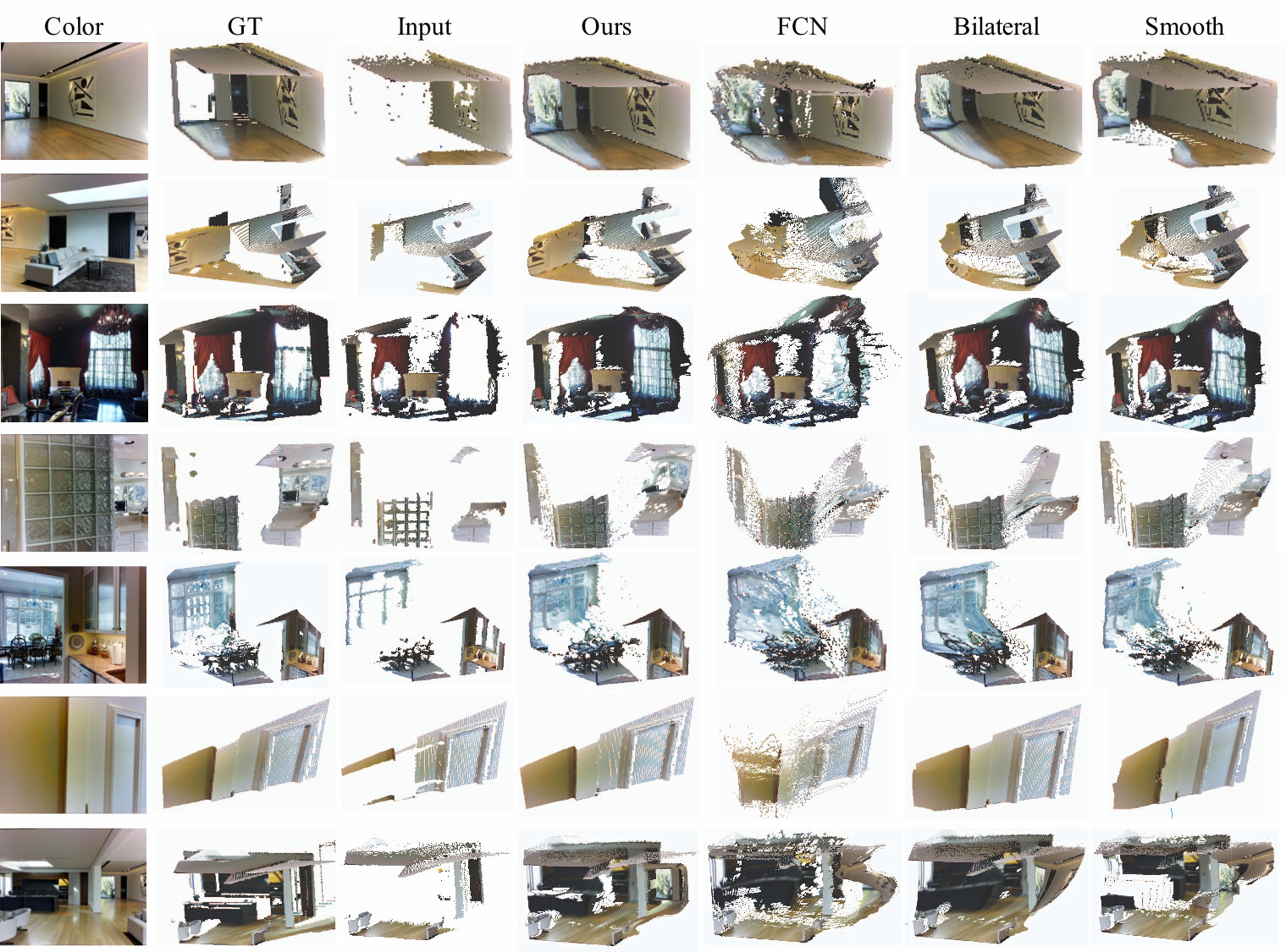}
\caption{{\bf Point cloud visualization of our method and other comparisons.} We convert the completed depth into point cloud. Our model produces better 3D geometry with fewer bleeding issue at the boundary.} 
\label{fig:pointcloud}
\end{figure*}

We also convert the completed depth maps into 3D point clouds for visualization and comparison, which are shown in Figure \ref{fig:pointcloud}.
The camera intrinsics provided in Matterport3D dataset is used to project each pixel on the depth map into a 3D point, and the color intensity are copied from the color image.
Each row shows one example, with the color image and point clouds converted from ground truth, input depth (i.e. the raw depth from sensor that contains a lot of missing area), and results of our method, FCN \cite{mao2016image}, joint bilateral filter \cite{silberman2012indoor}, and smooth inpainting.
Compared to other methods, our method maintains better 3D geometry and less bleeding on the boundary.

\end{appendices}

\end{document}